\documentclass{article} 
\usepackage{iclr2021_conference,times}

\usepackage{amsmath,amsfonts,bm,bbm}

\def\eqref#1{equation~\ref{#1}}

\def\1{\bm{1}}

\newcommand{\mean}{\mathrm{mean}}

\DeclareMathAlphabet{\mathsfit}{\encodingdefault}{\sfdefault}{m}{sl}
\SetMathAlphabet{\mathsfit}{bold}{\encodingdefault}{\sfdefault}{bx}{n}

\DeclareMathOperator*{\argmax}{arg\,max}
\DeclareMathOperator*{\argmin}{arg\,min}

\usepackage{xcolor}
\usepackage{xspace}
\usepackage{enumitem} 
\usepackage{hyperref}
\usepackage{url}
\usepackage{caption}
\usepackage{amssymb,graphicx,color}
\usepackage{wrapfig}
\usepackage{caption}
\usepackage{subcaption}
\usepackage{float}

\definecolor{myred}{rgb}{0.8,0,0}
\definecolor{mygreen}{rgb}{0,0.6,0}
\definecolor{myblue}{rgb}{0,0,0.7}
\definecolor{myorange}{rgb}{1.0, 0.55, 0.0}
\definecolor{mypink}{rgb}{1.0, 0.0, 0.4}

\newcommand{\mujoco}{{\sc MuJoCo}\xspace}

\definecolor{bluecite}{HTML}{0875b7}
\hypersetup{
  colorlinks=true,
  citecolor=bluecite,
  linkcolor=magenta
}

\title{Offline Reinforcement Learning Hands-On}

\author{Louis Monier\thanks{Equal contribution.}\\
        InstaDeep \\
        \texttt{louis.monier\thanks{@mines-paristech.fr}}\\
        \And
        Jakub Kmec\footnotemark[1]\\
        InstaDeep \\
        \texttt{j.kmec\thanks{@instadeep.com}}\\
        \And
        Alexandre Laterre\\
        InstaDeep \\
        \texttt{a.laterre\footnotemark[3]}\\
        \And 
        Thomas Pierrot\\
        InstaDeep \\
        \texttt{t.pierrot\footnotemark[3]}\\
        \And 
        Valentin Courgeau\\
        InstaDeep \\
        \texttt{v.courgeau\footnotemark[3]}\\
        \And
        Olivier Sigaud\\
        Sorbonne Universit\'{e} \\
        \texttt{olivier.sigaud\thanks{@upmc.fr}} \\
        \And
        Karim Beguir\\
        InstaDeep \\
        \texttt{kb\footnotemark[3]}\\
}

\iclrfinalcopy 
\begin{document}
\maketitle

\begin{abstract}

Offline Reinforcement Learning (RL) aims to turn large datasets into powerful decision-making engines without any online interactions with the environment. This great promise has motivated a large amount of research that hopes to replicate the success RL has experienced in simulation settings. This work ambitions to reflect upon these efforts from a practitioner viewpoint. We start by discussing the dataset properties that we hypothesise can characterise the type of offline methods that will be the most successful. We then verify these claims through a set of experiments and designed datasets generated from environments with both discrete and continuous action spaces. We experimentally validate that diversity and high-return examples in the data are crucial to the success of offline RL and show that behavioural cloning remains a strong contender compared to its contemporaries. Overall, this work stands as a tutorial to help people build their intuition on today’s offline RL methods and their applicability.


\end{abstract}

\section{Introduction}

Offline Reinforcement Learning holds the promise of bridging the gap between reinforcement learning algorithms and real-world applications. By taking advantage of large pre-collected datasets, it can mitigate the technical concerns associated with online data acquisition, as it is often expensive or dangerous to interact with real environments. These promises have triggered a surge of interest for research around this problematic and yielded notable improvements \citep{Kumar2020ConservativeQF, Wang2020CriticRR, Nair2020AcceleratingOR}. However, the fast publication pace ends up being confusing for non-experts, as it is not straightforward to know which offline algorithm should be used to address a particular application. Consequently, this 
might prevent its wide adoption, similar to what supervised learning has experienced with the availability of large datasets. Recent attempts were made to overcome this obstacle. \cite{levine2020offline} provides the reader with the conceptual tools needed to get started on research on offline RL. \cite{Fu2020D4RLDF} and \cite{Gulcehre2020RLUB} try to facilitate the measure of progress of research on offline RL by introducing benchmarks specifically designed for the offline setting. Despite these efforts, a hands-on up-to-date comparison of recent methods still ought to be carried out. The objective of this article is to provide the reader with a "cooking recipe" where we first focus on the ingredients, namely the offline methods and the dataset specifications. We then apply these algorithms on various datasets designed to highlight which method performs best in each setting. Finally, we discuss these findings so as to help the reader to build intuition on what might be the most appropriate method for their use case and start off the right foot.

\section{Background}

We consider a classic Reinforcement Learning (RL) framework with a Markov Decision Process (MDP), defined by the tuple ($\mathcal{S}$, $\mathcal{A}$, $p$, $r$, $\gamma$, $\rho_{0}$). $\mathcal{S}$ and $\mathcal{A}$ are the state and action spaces, respectively, and $\gamma \in (0, 1]$ is the discount factor. The dynamic or transition distribution is denoted as $p(s^{\prime} | s, a)$, the initial state distribution as $\rho_{0}(s)$, and the reward function as $r(s, a)$.
RL aims to maximize the expected sum of (discounted) rewards over trajectories $\tau$ by finding the optimal policy:
\begin{equation*}
    \pi^{*} = \argmax_{\pi} \mathbb{E}_{\tau \sim p_{\pi}(\tau)} \left[ \sum_{t=0}^{T} \gamma^{t} r(s_{t}, a_{t}) \right].
\end{equation*}

\paragraph{Distributional shift: the key difficulty in offline RL.} \label{sec:distShift}
RL algorithms, even the ones that can be trained from off-policy data, can not usually learn an optimal behaviour without additional on-policy interactions. This well-known issue, referred to as \textit{distributional shift} (also described as extrapolation/bootstrapping error accumulation \citep{Fujimoto2019OffPolicyDR, Kumar2019StabilizingOQ}) occurs when the used function approximator is trained under one distribution, i.e. the offline dataset, but evaluated on a different one -- the learned policy induces a different visited state-action distribution. Indeed, when entering states that are far outside the training distribution, even with strong inductive bases, we can hardly provide any convergence guarantee on the agent’s behaviour. Over the last few years, a considerable amount of algorithmic solutions has been published to address this problem. So far, the different approaches can be split into three different groups: explicit policy constraint methods, Q-values regularisation approaches and implicit policy constraint methods. 



\paragraph{Solution 1: Explicit Policy Constraint Methods.} A first solution to address the distributional shift is to constrain how much the learned policy might differ from the behaviour policy having created the dataset so that the distributional shift is bounded. Practically, one can force the parametrised policy being learned $\pi_{\phi}$ to take actions close to the behavioural distribution $\pi_{\beta}$ through a divergence measure $D_{m}$:
\begin{equation*}
    \argmax_{\phi} \mathbb{E}_{s \sim \mathcal{D}} \left[ \mathbb{E}_{a \sim \pi_{\phi}(. \mid s)} \left[ Q_{\theta}(s, a) \right] \right] \quad s.t. \quad D_m(\pi_{\phi}, \pi_{\beta}) \leq \epsilon.
\end{equation*}
This type of method fundamentally favors pessimism over risky exploration \citep{Fujimoto2019OffPolicyDR, Kumar2019StabilizingOQ}. This approach is mostly effective on expert demonstrations as it operates more like an imitation process rather than an offline algorithm \citep{Fujimoto2019BenchmarkingBD}. They also suffer from some limitations as (i) the estimated Q-values are often too conservative; (ii) the (unknown) behavioural policy $\pi_{\beta}$ has to be modeled.\\

Two categories of solutions appeared to overcome those limitations. We chose to focus on the latest, most promising and best representative methods from each, namely Critic Regularised Regression (CRR) \citep{Wang2020CriticRR, Nair2020AcceleratingOR} and Conservative Q-Learning (CQL) \citep{Kumar2020ConservativeQF}. Neither of the two require modelling the prior policy, nor enforce an explicit constraint to avoid out-of-distribution actions.
    
\paragraph{Solution 2: Q-values Regularisation.}
    The CQL algorithm inserts an additional regularisation term on top of standard policy evaluation steps to learn a conservative Q-function and avoids overestimation issues, highly detrimental when boostrapping:
    \begin{equation} \label{eq:CQLloss}
        \argmin_{\theta} \:\: \alpha \mathbb{E}_{s \sim \mathcal{D}} \left[ \log{\sum_{a} \exp{ Q_{\theta}(s, a)} } - \mathbb{E}_{a \sim \pi_{\beta}(. \mid s)} \left[ Q_{\theta}(s, a) \right] \right] + \dfrac{1}{2} \mathbb{E}_{(s, a, s^{\prime}) \sim \mathcal{D}} \left[ (Q_{\theta} - \mathcal{B}^{\pi_{\phi}} Q)^{2} \right] 
    \end{equation}

    Intuitively, the \textit{LogSumExp} term squashes down the Q-values, especially the overestimated ones. The second term compensates the "push down" effect by maximizing the Q-values of (state-action) pairs sampled from the dataset. The resulting modified objective prevents overestimation without constraining the policy to stay close the behavioural distribution. This new critic update yields strong theoretical guarantees: the expected Q-value under $\pi$ is actually a lower bound of its true value. In Section~\ref{sec:ExperimentsResults}, we provide further practical implementation remarks for CQL.

\paragraph{Solution 3: Implicit Policy Constraint Methods.} 
    Another approach is to implicitly apply a constraint on the policy to dissuade it from selecting out-of-distribution actions. The idea is to create a weighted behavioural cloning objective where bad actions are discarded and good ones are used to train the agent. In practice, the policy update re-weights the state-action pairs from the buffer leveraging the advantage estimations coming from the critic:
    \begin{equation*}
        \argmax_{\phi} \mathbb{E}_{(s, a) \sim \mathcal{D}} \left[ \log \left( \pi_{\phi}(a \mid s) \right) f(Q_{\theta}, \pi_{\phi}, s, a) \right],
    \end{equation*}
    where the so-called \emph{filtered} function pushes the policy towards the most reward-promising transitions sampled from the buffer. Note that in the past years, several methods tried to implicitly induce constraints in the RL objective \citep{Ghasemipour2020EMaQEQ, Siegel2020KeepDW, Wu2019BehaviorRO}. We focus on the latest up-and-coming method CRR which was published approximately at the same time as AWAC \citep{Nair2020AcceleratingOR}. Both algorithms implement the implicit constraint similarly. See Appendix~\ref{app:baselines} for the definitions used in practice for the \emph{filtered} function and the advantage.

\section{Dataset characteristics}
In this section, we explore the main characteristics of offline RL datasets. The experimental dataset setup is further detailed in Appendix \ref{section:dataset-creation-protocol} for our experiments presented in Section~\ref{sec:ExploringDatasetEdgeCases} and the characteristics of each dataset are presented in Appendix~\ref{sec:DatasetVis}.

Online RL is effectively a feedback loop: the chosen actions determine the training data. In offline RL, the dataset and amount of exploratory behaviour are kept fixed. We are therefore limited to finding the best policy for the MDP defined by the dataset rather than solving the true MDP that exists in the real world. Therefore, having a limited dataset is detrimental to the performance of an offline agent.

\textbf{Trajectories quality vs state-action coverage?} This trade-off underpins offline RL analogously to the exploration vs.\ exploitation dilemma in online RL. On one hand, the dataset must include \emph{high-quality} actions i.e.\ leading to high-reward episodes. On the other hand, learning a policy that outperforms the behavioural policy requires the dataset to include "\textit{bad}" (exploratory) actions that the behavioural policy would not have taken. That is, a dataset with a higher state-action \emph{coverage}. See the first experiment in Section~\ref{sec:ExploringDatasetEdgeCases} as well as Figures \ref{fig:expertEps0}--\ref{fig:expertEps1}, App.~\ref{sec:DatasetVis} and the interpretation therein.

At one extreme lies a \textbf{fully-random} behavioural policy which collects a dataset with very good action space coverage but only for the states that are reachable by such policy. Therefore, in environments requiring elaborate exploration strategies, the dataset might not contain enough high-reward transitions to learn a satisfactory policy. Furthermore, datasets collected by random policies are not very common in practice.

At the other end is a \textbf{fully-deterministic} behavioural policy, e.g.\ logs generated by a deterministic heuristic. It is very difficult to learn policies from offline data that outpreform such heuristics: each state is only paired with a single action leading to a poor action-state coverage. Current approaches rely on the generalization, or extrapolation capability of the function approximation. This provides few guarantees and usually leads to overestimation errors that accumulate (\textit{distributional shift}). The field of Causal Inference provides a more extensive treatment of such scenarios with more underlying assumptions that provide additional guarantees. It introduces the concept of \textit{counterfactual queries}, which attempt to estimate the value---or so-called \emph{effect}---of unobserved actions \citep{pearl2009causality}.

To summarise, neither a fully random nor a fully deterministic behavioural policy are suitable for collecting offline RL datasets. There are two key properties the behavioural policy should have to obtain an actionable dataset: 1) \emph{Sufficient quality} to reach all high-reward regions (Fig.\ \ref{fig:DirectComparison3}). 2) \emph{Sufficient coverage} to allow for fruitful exploration (Section~\ref{sec:ExploringDatasetEdgeCases} on outperforming the dataset). 

This being said, quantifying the level of exploration in an offline dataset is a difficult problem in itself. Proxies can be used to better understand what type of behaviour is available within a dataset. Some examples include: (i) reward distribution, (ii) actions distribution (e.g.\ policy entropy), (iii) episode length distribution, (iv) maximum reward in dataset / maximum attainable reward, (v) state coverage, (vi) state-action coverage. We plot the first five metrics for each of the datasets used in our experiments in Appendix~\ref{sec:DatasetVis}. It is important to note that visualizing the last two metrics is not always trivial in high dimensional environments.



\section{Experiments and Results}
\label{sec:ExperimentsResults}

The goal of this section is to help the reader better understand how current state-of-the-art methods (CQL and CRR) perform as a function of the dataset quality and coverage ; we also highlight some common failure modes. We first show how the chosen offline algorithms perform in a simple interpretable grid world environment with discrete actions. We then run the same experiments on more realistic tasks which include higher dimensional environments with continuous action spaces. We also compare CQL and CRR approaches with naive off-policy baseline algorithms (DQN/SAC) and Behaviour Cloning (BC). Baseline implementation details can be found in Appendix~\ref{app:baselines}.

\subsection{Exploring dataset edge cases}
\label{sec:ExploringDatasetEdgeCases}
In this section, we run three experiments on a simple MiniGrid environment \citep{gym_minigrid} to help the reader develop a better understanding of the core issues specific to offline RL. The goal is to strip away any confounding factor such as the need perception in pixel observations or a stochasticity in the environment that could make the results more convoluted. 

\paragraph{How much does performance vary with dataset quality?} 
As mentioned in the previous section, dataset quality can be described in many ways. Here, we use an $\epsilon$-greedy expert with 3 different values of $\epsilon$ as described in Appendix~\ref{app:discreteProtocol}. As $\epsilon$ gets larger, the action selection becomes more stochastic and deviates from the expert baseline.
\begin{figure}[h!]
    \centering
    \includegraphics[width=\textwidth]{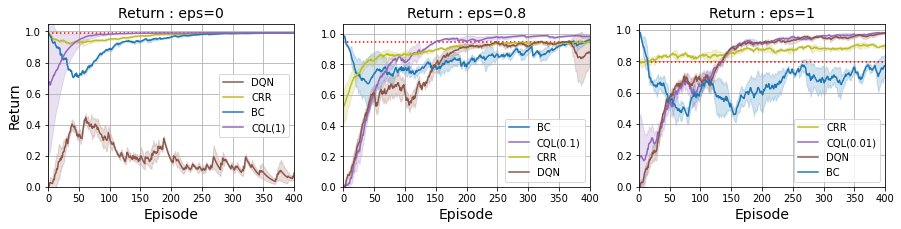} 
    \caption{Comparison of CQL, CRR (with exponential filter), DQN and BC on datasets of different quality. The red dotted line shows the average episode return for each dataset ($0.991$, $0.947$, $0.796$ respectively).}
    \label{fig:DirectComparison3}
\end{figure}

We carried out experiments across datasets of varying quality and a grid search over the key hyper-parameters for each algorithm. All the results can be found in Appendix~\ref{sec:MisResults}. Figure~\ref{fig:DirectComparison3}
depicts the results obtained when training each agent on an expert, medium-quality and fully random dataset, respectively. CQL consistently achieves the highest returns. However, this comes at the cost of a high sensitivity to the value of alpha which had to be tuned precisely for each dataset. CQL with small alpha ($\alpha=0.01$) performs better on datasets that contain more randomness, but fails on expert data (see Fig. \ref{fig:CQLexperiment}). The opposite is true for large values of alpha, i.e.\ for $\alpha=1$. This is expected, since alpha determines how much weight is put on the regularising term (Eq.\ \ref{eq:CQLloss}) which implicitly constrains the policy to lie close to the dataset distribution. CRR$_{\exp}$ (CRR with an exponential filter) achieves lower returns but is more robust to the choice of hyper-parameters, such that the same value performs well across all experiments. Finally, the baselines also behave as expected, the performance of BC is upper-bounded by the average return of the trajectories in each dataset: only preforms well on the expert dataset. DQN, on the other hand, fails in the expert settings due to distributional shift (discussed in sec. \ref{sec:distShift}), but performs well on a fully random dataset thanks to its high state-action space coverage.

\paragraph{How much can we improve upon the dataset performance?} To test this, we construct a highly stochastic dataset with very sparse rewards. It is essential to employ RL methods in such settings, since other approaches, such as behavioural cloning, are unable to exploit the sequential nature of the problem. We collect a dataset by a random agent acting in the Lava environment shown in Figure \ref{fig:expertPathsLava}.
\begin{wrapfigure}{r}{0.5\textwidth}
    \begin{center}
        \includegraphics[width=0.5\textwidth]{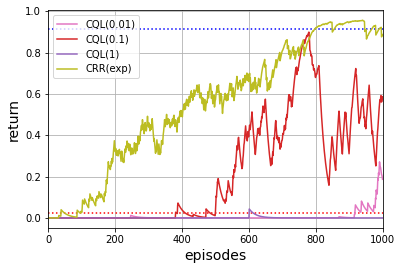}
    \end{center}
    \captionsetup{justification=justified,margin=0cm, belowskip=-5pt}
    \vspace{-0.5cm}
    \caption{Red and blue dotted lines show the average and maximum episode return respectively.}
    \label{fig:lavaExperiment}
\end{wrapfigure}
 A detailed visualization of the dataset characteristics can be found in Figure \ref{fig:randomLavaVis}, Appendix \ref{sec:DatasetVis}. Note that it only includes 3.1\% of positive trajectories with an average episode return of 0.024. Figure \ref{fig:lavaExperiment} on the right shows the performance of our selected methods on this highly sparse dataset. We can see that CRR$_{\exp}$ exceeds the maximum episode return in the dataset (blue dotted line), which means it must piece together parts of different trajectories. This behaviour has been previously referred to as stitching \citep{Fu2020D4RLDF}. CQL, with the right alpha still greatly improves upon the average dataset behaviour, but does not exceed the maximum. BC and DQN fail in this settings which is why we didn't include them.

\paragraph{Can existing methods extract optimal policies from multi-modal datasets?}
\begin{wraptable}{r}{0pt}
    \begin{tabular}{ c | c | c } 
        \hline
        \textbf{Method} & \textbf{Hyperparameters} & \textbf{Ep. length}  \\ 
        \hline
        CQL & $\alpha=0.001$ & 500 \\ 
        & $\alpha=0.01$ & 13 \\
        & $\alpha=0.1$, $\alpha=1$ & 17 \\
        \hline
        CRR$_{\exp}$ & $\beta=1$ & 17 \\
        & $\beta=0.01$ & 500 \\
        \hline
        CCRR & $\alpha = 0.01, \beta=1$ & 17 \\
         & $\alpha = 0.01, \beta=0.01$ & 13 \\
        \hline
        BC & & 17 \\
        \hline
    \end{tabular}
    \captionsetup{justification=justified,margin=0cm}
    \caption{Final episode length on the multi-modal dataset described in Appendix~\ref{app:discreteProtocol}.} 
    \label{tab:multiModalResults}
\end{wraptable} 

The experiment described in Appendix~\ref{app:discreteProtocol} aims to test the agents ability to recover the optimal behaviour from a dataset containing trajectories of policies of various qualities.
The most straightforward way to tackle out-of-distribution actions is to constrain the policy to lie close to the behaviour policy. Methods that use such approach e.g. CQL, are expected to fail in this setting. Indeed, we can see in Table \ref{tab:multiModalResults}  that CQL with a high value of alpha converges to the incorrect sub-optimal policy. On the other hand, if we decrease alpha too much, it fails to learn at all and suffers from the distributional shift. There is a value of alpha for which CQL converges to the correct policy, but the algorithm is highly sensitive and was difficult to tune correctly. BC and CRR with an exponential filter fail to learn the correct policy. We tested a variant of CRR by incorporating a conservative critic that uses the CQL penalty (CCRR). The conservative critic made the advantage estimates more accurate and successfully learned the optimal policy, but introduced the additional complexity of tuning the value of alpha. \\

From these experiments, practitioners might want to keep in mind that the quality of the offline dataset is crucial to train efficient decision-making agents, even more than in the supervised setting. In scenarios where the dataset contains highly stochastic behaviours, it is worth trying an off-policy online method such as DQN, which directly optimizes the true objective without imposing any constraints. In most practical scenarios however, the dataset will have gaps in its state-action coverage and online methods will fail due to distributional shift. This is where CQL and CRR prove useful. Properly tuned CQL usually yields the best results, but CRR is more robust to hyper-parameter selection and works better on datasets with sparse rewards. Lastly, if we have a dataset that was collected by a highly deterministic policy, BC is going to be very difficult to outperform. Understanding how these methods perform on environments with increasing complexity is what we deal with in the next section.
    
\subsection{Making the most out of offline RL for continuous control}
\label{sec:continuousExp}

We present here experiments on challenging continuous control tasks for people looking to leverage their dataset in more realistic settings. Similarly to the previous experiments with discrete action spaces, we created three generic datasets to cover a wide range of real-world scenarios while keeping it as simple as possible: high quality data (expert policy behaviour), medium quality (average performance) and purely random data. We intentionally selected environments with an increasing complexity to evaluate the algorithms in various contexts: a simple 2D grid (\texttt{PointMaze-v0}) and two classic \mujoco benchmark tasks (\texttt{HalfCheetah-v2} and \texttt{Humanoid-v2}). Additional details on the datasets creation protocol and the environments can be found in Appendix~\ref{app:continuousProtocol} and \ref{app:environments} respectively. The experiments below depict how the offline methods actually perform when they are implemented following the instructions from the papers, as well as when using the suggested hyper-parameter values.

\paragraph{What is the best offline training strategy when dealing with continuous control tasks?}

\begin{figure}[h!]
    \centering
    \includegraphics[width=0.99\textwidth]{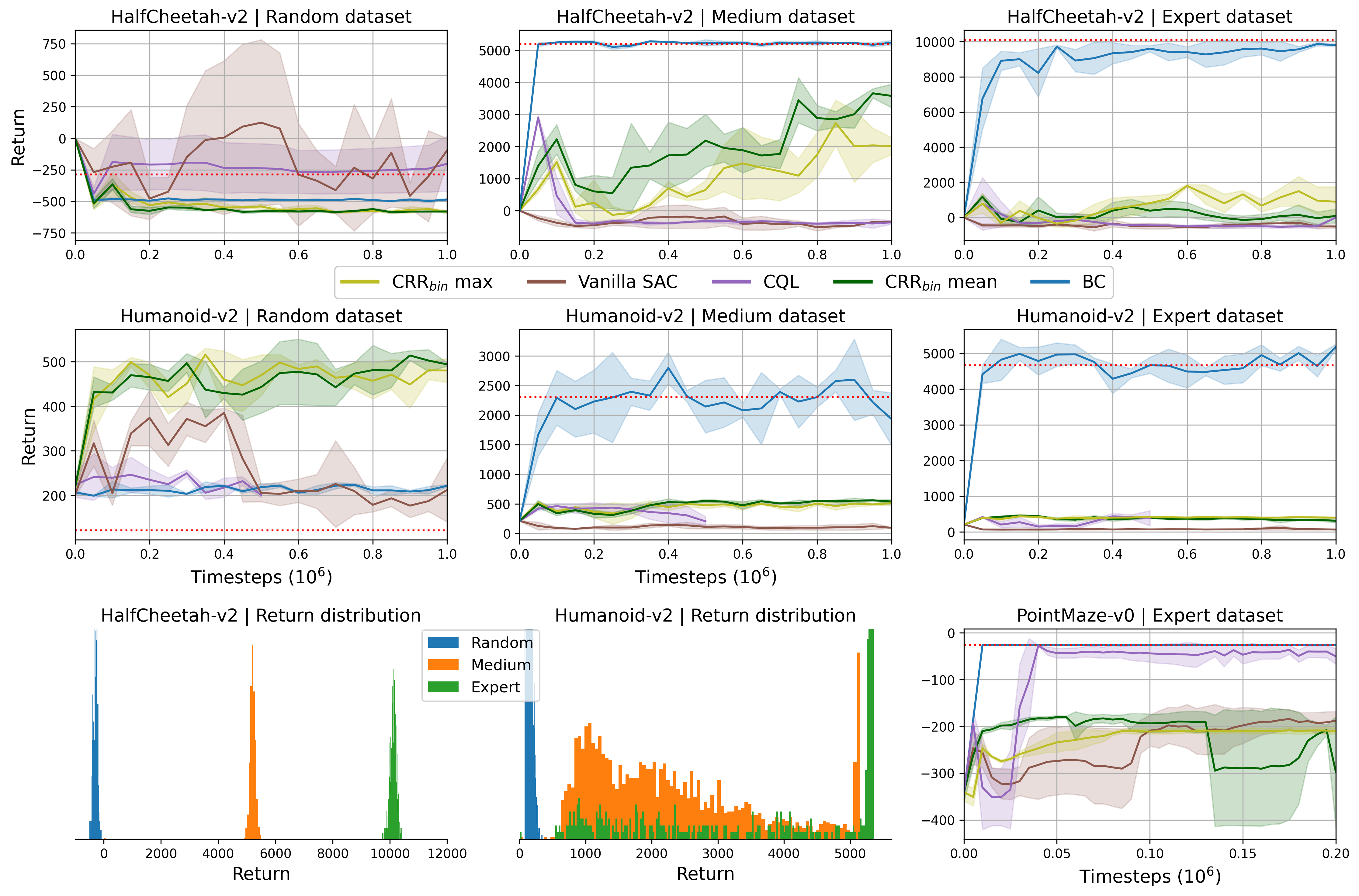}
    \captionsetup{justification=justified,margin=0cm}
    \caption{Comparison of CQL, CRR, SAC and BC on different environments and datasets of various qualities. The red dotted line indicates the average episode return and the normalized return distributions are plotted in the bottom left for each dataset.}
    \label{fig:ContinuousExp}
\end{figure}

The experiments depicted in Figure~\ref{fig:ContinuousExp} first show that naive application of an off-policy algorithm like SAC fails when applied in the offline setting. As for BC, it reaches a near-optimal strategy with medium and high-quality data but automatically fails when a large part of random or low-performing behaviours is present in the dataset.

As shown in Figure~\ref{fig:ContinuousExp} top-center, CRR is the only offline method able to partly retrieve the performance contained in the medium quality dataset on \texttt{HalfCheetah-v2} but does not improve upon it. By definition, CRR is only able to mimic a subset of good action decisions. Indeed, the agent is implicitly constrained to lie in the vicinity of the best performing trajectory in the static dataset through the use of the \emph{filtered} function. Therefore, it cannot extract information across trajectories and therefore cannot outperform the dataset performance. Note that it fails when the action distribution is narrow or when its entropy is high (no useful information to extract). In practice, the exponential \emph{filtered} function can cause numerical instabilities leading to crashes during the training. Therefore, we mainly use the binary function (CRR$_{bin}$) with both \emph{mean} and \emph{max} advantage estimates. 

As far as CQL is concerned, it remains unstable and is unable to learn robustly across all the dataset-task pairs. Moreover, we found it difficult to implement: the authors recommend using the \textit{LogSumExp} function provided by Torch or TensorFlow libraries for the discrete actions setting and suggest otherwise for the continuous case, while the official repository seems to use it in either case. They also use an additional term in their code for the Q-value regulariser:
$\sum_{a^{i}_{t+1} \sim \pi (\textbf{a}_{t+1} \mid s_{t+1})}^{N} \exp\left\{Q(s_{t+1}, a^{i}_{t+1}) - \log \pi (a^{i}_{t+1} \mid s_{t+1})\right\}$, 
which is not mentioned in the paper. Intuitively, they likely include these additional samples (from the region of interest where we might find higher Q-values) to improve the accuracy of the regularisation term\footnote{After contacting the authors, they confirmed that all the available sample sources are needed to evaluate the regulariser more accurately. They also suggest a higher number of samples ($N \approx$ 50--100) instead of the paper-recommended value ($N \approx$ 10).}. Finally, they also use a temperature hyper-parameter to make the soft-maximum estimate more accurate and remove the non-deterministic part of the Q-value target (additional entropy term) which are not mentioned in the paper. Overall, applying the CQL method is appealing but even after fixing all these discrepancies, we were not able to replicate the results. We believe the fragile hyper-parameters equilibrium makes practical implementations difficult and subject to high variance. 
Finally, in a simpler continuous environment like \texttt{PointMaze-v0}, CQL and BC interestingly perform best and retrieve the expert behaviour whereas CRR gets stuck in a local minimum. 

Overall, offline RL is undeniably a tough problem. The takeaway is that spending a lot of resources on data collection and/or algorithm customization does not guarantee a robust strong performance for complex continuous control tasks as the latest methods are unable to generalize well on most new dataset-task pairs. A safe bet is BC which remains a strong and robust baseline for any continuous settings as long as the dataset contains qualitative data. We provide further fine-tuning experiments in Appendix~\ref{sec:MisResults} to evaluate whether it is possible to get further gains of performance with a limited additional exploration after offline training when a simulator can be used.

\section{Conclusion and Discussion}

In this paper, after presenting three methods from recent offline RL contributions, we take a step back from the literature to return to practical considerations: data and algorithms. For the former, we discuss various dataset properties that may play a role in a successful deployment of offline RL. Both the quality of the trajectories therein contained and their coverage of the state-action space offer a key trade-off when it comes to data. We devise a collection of experiments to document their interface through the lens of three algorithms (BC, CRR and CQL) and three dataset quality levels (random, medium and expert). Since the problems we tackle are undocumented, we provide foundations to help the reader better grasp the issues and suitability of different offline RL methods. We hypothesise that datasets that are of interest in offline RL are generated by medium quality policies, ideally with high stochasticity which lead to a dataset in which we can expect to be able to outperform the data generating process. Overall, BC strengthens its position as baseline with a robust performance across datasets with qualitative data by matching the near-optimal strategy contained in the buffer. On discrete action spaces, CQL exhibits good recovery properties for medium quality data and CRR occasionally outperforms the dataset baseline. Furthermore, when testing the methods on more complex environments, we found it even harder to beat the baselines. Even though CQL was shown to preform well on very complex environments, it can be difficult and expensive to tune because of the fragile hyper-parameter equilibrium. 

The selection of the offline RL method to use and its success cannot be made without the aforementioned considerations, that is, no method performs uniformly better independently of the use case. However, dealing with custom datasets uncovers many practical pitfalls that have not been addressed in the literature such as algorithms' fitness for different levels of data quality. Given the pace at which the field of offline RL is evolving, we expect it to become a central question in the near future.

\newpage

\bibliography{iclr2021_conference}

\begin{thebibliography}{18}
\providecommand{\natexlab}[1]{#1}
\providecommand{\url}[1]{\texttt{#1}}
\expandafter\ifx\csname urlstyle\endcsname\relax
  \providecommand{\doi}[1]{doi: #1}\else
  \providecommand{\doi}{doi: \begingroup \urlstyle{rm}\Url}\fi

\bibitem[Chevalier-Boisvert et~al.(2018)Chevalier-Boisvert, Willems, and
  Pal]{gym_minigrid}
Maxime Chevalier-Boisvert, Lucas Willems, and Suman Pal.
\newblock Minimalistic gridworld environment for openai gym.
\newblock \url{https://github.com/maximecb/gym-minigrid}, 2018.

\bibitem[Cideron et~al.(2020)Cideron, Pierrot, Perrin, Beguir, and
  Sigaud]{Cideron2020QDRLEM}
Geoffrey Cideron, Thomas Pierrot, N.~Perrin, Karim Beguir, and Olivier Sigaud.
\newblock Qd-rl: Efficient mixing of quality and diversity in reinforcement
  learning.
\newblock \emph{ArXiv}, abs/2006.08505, 2020.

\bibitem[Fu et~al.(2020)Fu, Kumar, Nachum, Tucker, and Levine]{Fu2020D4RLDF}
Justin Fu, Aviral Kumar, Ofir Nachum, G.~Tucker, and Sergey Levine.
\newblock D4rl: Datasets for deep data-driven reinforcement learning.
\newblock \emph{ArXiv}, abs/2004.07219, 2020.

\bibitem[Fujimoto et~al.(2018)Fujimoto, Hoof, and
  Meger]{Fujimoto2018AddressingFA}
Scott Fujimoto, H.~V. Hoof, and David Meger.
\newblock Addressing function approximation error in actor-critic methods.
\newblock \emph{ArXiv}, abs/1802.09477, 2018.

\bibitem[Fujimoto et~al.(2019{\natexlab{a}})Fujimoto, Conti, Ghavamzadeh, and
  Pineau]{Fujimoto2019BenchmarkingBD}
Scott Fujimoto, Edoardo Conti, Mohammad Ghavamzadeh, and Joelle Pineau.
\newblock Benchmarking batch deep reinforcement learning algorithms.
\newblock \emph{ArXiv}, abs/1910.01708, 2019{\natexlab{a}}.

\bibitem[Fujimoto et~al.(2019{\natexlab{b}})Fujimoto, Meger, and
  Precup]{Fujimoto2019OffPolicyDR}
Scott Fujimoto, D.~Meger, and Doina Precup.
\newblock Off-policy deep reinforcement learning without exploration.
\newblock In \emph{ICML}, 2019{\natexlab{b}}.

\bibitem[Ghasemipour et~al.(2020)Ghasemipour, Schuurmans, and
  Gu]{Ghasemipour2020EMaQEQ}
Seyed Kamyar~Seyed Ghasemipour, Dale Schuurmans, and Shixiang Gu.
\newblock Emaq: Expected-max q-learning operator for simple yet effective
  offline and online rl.
\newblock \emph{ArXiv}, abs/2007.11091, 2020.

\bibitem[Gulcehre et~al.(2020)Gulcehre, Wang, Novikov, Paine, Colmenarejo,
  Zolna, Agarwal, Merel, Mankowitz, Paduraru, Dulac-Arnold, Li, Norouzi,
  Hoffman, Nachum, Tucker, Heess, and Freitas]{Gulcehre2020RLUB}
Caglar Gulcehre, Ziyu Wang, A.~Novikov, T.~L. Paine, Sergio~Gomez Colmenarejo,
  Konrad Zolna, Rishabh Agarwal, Josh Merel, Daniel~J. Mankowitz, Cosmin
  Paduraru, Gabriel Dulac-Arnold, J.~Li, Mohammad Norouzi, Matt Hoffman, Ofir
  Nachum, G.~Tucker, Nicolas Heess, and N.~D. Freitas.
\newblock Rl unplugged: Benchmarks for offline reinforcement learning.
\newblock \emph{ArXiv}, abs/2006.13888, 2020.

\bibitem[Haarnoja et~al.(2018)Haarnoja, Zhou, Abbeel, and
  Levine]{Haarnoja2018SoftAO}
T.~Haarnoja, Aurick Zhou, P.~Abbeel, and S.~Levine.
\newblock Soft actor-critic: Off-policy maximum entropy deep reinforcement
  learning with a stochastic actor.
\newblock In \emph{ICML}, 2018.

\bibitem[Kumar et~al.(2019)Kumar, Fu, Tucker, and
  Levine]{Kumar2019StabilizingOQ}
A.~Kumar, Justin Fu, G.~Tucker, and S.~Levine.
\newblock Stabilizing off-policy q-learning via bootstrapping error reduction.
\newblock \emph{ArXiv}, abs/1906.00949, 2019.

\bibitem[Kumar et~al.(2020)Kumar, Zhou, Tucker, and
  Levine]{Kumar2020ConservativeQF}
Aviral Kumar, Aurick Zhou, G.~Tucker, and Sergey Levine.
\newblock Conservative q-learning for offline reinforcement learning.
\newblock \emph{ArXiv}, abs/2006.04779, 2020.

\bibitem[Levine et~al.(2020)Levine, Kumar, Tucker, and Fu]{levine2020offline}
Sergey Levine, Aviral Kumar, George Tucker, and Justin Fu.
\newblock Offline reinforcement learning: Tutorial, review, and perspectives on
  open problems.
\newblock \emph{arXiv preprint arXiv:2005.01643}, 2020.

\bibitem[Mnih et~al.(2013)Mnih, Kavukcuoglu, Silver, Graves, Antonoglou,
  Wierstra, and Riedmiller]{Mnih2013PlayingAW}
V.~Mnih, K.~Kavukcuoglu, D.~Silver, A.~Graves, Ioannis Antonoglou, Daan
  Wierstra, and Martin~A. Riedmiller.
\newblock Playing atari with deep reinforcement learning.
\newblock \emph{ArXiv}, abs/1312.5602, 2013.

\bibitem[Nair et~al.(2020)Nair, Dalal, Gupta, and
  Levine]{Nair2020AcceleratingOR}
Ashvin Nair, Murtaza Dalal, Abhishek Gupta, and Sergey Levine.
\newblock Accelerating online reinforcement learning with offline datasets.
\newblock \emph{ArXiv}, abs/2006.09359, 2020.

\bibitem[Pearl(2009)]{pearl2009causality}
Judea Pearl.
\newblock \emph{Causality}.
\newblock Cambridge university press, 2009.

\bibitem[Siegel et~al.(2020)Siegel, Springenberg, Berkenkamp, Abdolmaleki,
  Neunert, Lampe, Hafner, and Riedmiller]{Siegel2020KeepDW}
Noah Siegel, Jost~Tobias Springenberg, Felix Berkenkamp, Abbas Abdolmaleki,
  Michael Neunert, T.~Lampe, Roland Hafner, and Martin~A. Riedmiller.
\newblock Keep doing what worked: Behavioral modelling priors for offline
  reinforcement learning.
\newblock \emph{ArXiv}, abs/2002.08396, 2020.

\bibitem[Wang et~al.(2020)Wang, Novikov, Zolna, Springenberg, Reed, Shahriari,
  Siegel, Merel, Gulcehre, Heess, and Freitas]{Wang2020CriticRR}
Ziyu Wang, A.~Novikov, Konrad Zolna, Jost~Tobias Springenberg, Scott Reed,
  B.~Shahriari, N.~Siegel, Josh Merel, Caglar Gulcehre, Nicolas Heess, and
  N.~D. Freitas.
\newblock Critic regularized regression.
\newblock \emph{ArXiv}, abs/2006.15134, 2020.

\bibitem[Wu et~al.(2019)Wu, Tucker, and Nachum]{Wu2019BehaviorRO}
Y.~Wu, G.~Tucker, and Ofir Nachum.
\newblock Behavior regularized offline reinforcement learning.
\newblock \emph{ArXiv}, abs/1911.11361, 2019.

\end{thebibliography}
\bibliographystyle{iclr2021_conference}

\newpage

\appendix
\section{Specs and protocols}

\subsection{Dataset Creation Protocol}
\label{section:dataset-creation-protocol}
\paragraph{Discrete action spaces} \label{app:discreteProtocol} We used a simple double-DQN to train an expert agent on each environment. The fully trained expert is then used to collect datasets for each experiment.

\textit{EXPERIMENT 1}: we use an $\epsilon$-greedy expert policy with increasing $\epsilon$ values to interact with a simple empty 6x6 MiniGrid environment \citep{gym_minigrid}. The state in this environemnt is defined by cell position and orientation and there are 3 actions: turn left/right and move forward. The dataset used in the experiment are also visualized in Appendix \ref{sec:DatasetVis}.

\begin{figure}[H]
    \begin{subfigure}[b]{0.43\textwidth}
        \centering
        \includegraphics[width=\textwidth]{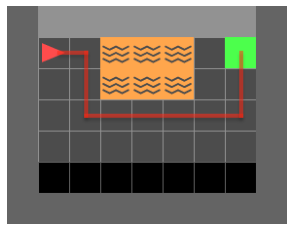}
        \caption{Expert trained with greedy Q-learning updates to learn the shortest path to the goal (13 steps).}
        \label{fig:expertPathsLava:a}
    \end{subfigure}
    \hfill
    \begin{subfigure}[b]{0.43\textwidth}
        \centering
        \includegraphics[width=\textwidth]{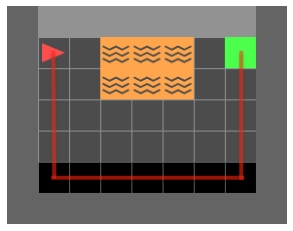}
        \caption{Expert trained with SARSA updates with high epsilon to converge to a safe path (17 steps).}
        \label{fig:expertPathsLava:b}
    \end{subfigure}
    \captionsetup{justification=justified,margin=0cm}
    \caption{The two expert trajectories used for data collection.}
    \label{fig:expertPathsLava}
\end{figure}

\textit{EXPERIMENT 2}: we use a fully random agent to collect data in the Lava environment shown in Figure \ref{fig:expertPathsLava}. Most random walks in this environment end up in Lava which terminates the episode with zero reward. This results in a dataset which is very sparse. 

\textit{EXPERIMENT 3}: the multi-modal dataset used in the third experiment contains data collected by agents with two different policies.

To collect the final dataset, we generate 20\% of the trajectories using expert A and the remaining 80\% of the trajectories using expert B. Both experts are epsilon-greedy with $\epsilon = 0.1$. As usual, the final dataset can be visualized in Appendix \ref{sec:DatasetVis}, see Figure \ref{fig:MultiModalEVis}. The goal is to create a dataset that contains a majority of relatively high reward, but sub-optimal trajectories, with a smaller proportion of optimal trajectories.

\paragraph{Continuous action spaces} \label{app:continuousProtocol}
We used QD-RL \citep{Cideron2020QDRLEM}, TD3 \citep{Fujimoto2018AddressingFA} and SAC \citep{Haarnoja2018SoftAO} to generate homemade offline datasets on \texttt{PointMaze-v0}, \texttt{HalfCheetah-v2} and \texttt{Humanoid-v2} benchmarks respectively. Each dataset contains $10^6$ transitions. We chose three generic settings in which most of real-world situations can be classified:
\begin{itemize}[leftmargin=*]
    \item \underline{"Random":} \\
    To collect the dataset, we simply let a random agent wandering around in the environment. Allegedly, random datasets show the largest state-action coverage even though it is not enough to ensure deep exploration. In a realistic setting, it corresponds to situations where the logged data is meaningless but still represents a possible physical state of the system under consideration. For example, let's say we want to train a decision making agent to optimize a power plant operational efficiency. Arbitrary data does not feature any relevant operation strategy, yet it outlines possible values. 
    
    \item \underline{"Medium":} \\
    The RL agent is trained online until an average return of approximately 5000 is reached on \texttt{HalfCheetah-v2} and 2500 on \texttt{Humanoid-v2}. Then, the transitions are saved. This setting is the most realistic as it represents a wide range of possible real-word scenarios where datasets with some sub-optimal strategies can be easily collected (hand-engineered policies or human demonstrations for instance). Going back to the power plant example, let's consider that human operators handle the energy production. Logs of the relevant plant parameters are collected and may reflect poor production performance every now and then (non-expert policy). 
    
    In general, medium quality datasets are also interesting to investigate because this setting makes it ultimately easier to detect potential improvements of an offline agent upon the best performing trajectory of the static data.
    
    \item \underline{"Expert":} \\
    Same data acquisition protocol as "medium" but with a return around 10000 on \texttt{HalfCheetah-v2}, 5000 on \texttt{Humanoid-v2} and -26 on \texttt{PointMaze-v0}. This setting highlights near-expert behaviours. A successful offline RL agent is expected to reach at least the same level of performance as contained in the buffer. But mostly it should be able to generalize to new situations and perform similarly. 
\end{itemize}

The histograms of the return distributions are represented at the bottom left of Figure~\ref{fig:ContinuousExp}. The return profiles for \texttt{HalfCheetah-v2} show sharp and narrow distributions. For \texttt{Humanoid-v2}, the distribution contains multi-modal transitions because of the larger return coverage in the medium and expert settings.

\subsection{Baselines Implementation Details}
\label{app:baselines}


\begin{itemize}[leftmargin=*]
    \item \textbf{Naive off-policy algorithms in the offline setting:} \\
    As a reference baseline, we naively apply off-the-shelf off-policy RL algorithm in the offline setting (DQN \citep{Mnih2013PlayingAW} and SAC \citep{Haarnoja2018SoftAO} for discrete and continuous control tasks respectively) to show its failure for every dataset-algorithm-task configurations caused by the distributional shift problem.
    \item \textbf{Behavioural Cloning (BC):} \\
    This is the simplest form of imitation learning as it does not require environment interactions. In practice, it is the easiest method to implement as it just copies/mimics actions from the batch, so no RL process is involved. BC only controls in-distribution errors and does not address the out-of-distribution actions problem: \\
    \begin{equation*}
        \argmin_{\phi} \mathbb{E}_{(s, a) \sim \mathcal{D}} \left[ \mathbb{E}_{a^{\prime} \sim \pi_{\phi}(. \mid s)} \left[ D_{m} \left( a^{\prime}, a \right) \right] \right]
    \end{equation*}
    where $D_{m}$ is a divergence measure. We use the cross-entropy and the mean square error loss for discrete and continuous control domains respectively.
    \item \textbf{Critic Regularised Regression (CRR):} \\
    The CRR algorithm acts more like a smart BC method as it cannot pull information across the best performing trajectory in the buffer. The \emph{filtered} function which exploits the best actions and discards the bad ones has several definitions in practice:
    $$f(Q_{\theta}, \pi_{\phi}, s, a) = \mathbbm{1} \left[ A^{\pi_{\phi}}(s, a) > 0 \right] \text{  or  }\exp \left( A^{\pi_{\phi}}(s, a) / \lambda \right).$$
    Moreover, CRR mainly uses two definitions for the advantage $A^{\pi}$. 
    \begin{itemize}
        \item The \emph{mean} advantage: $\hat{A}_{\mean}(s_{t}, a_{t}) = Q_{\theta}(s_{t}, a_{t}) - m^{-1} \sum_{j = 1}^{m} Q_{\theta}(s_{t}, a_{j})$.
        \item the \emph{max} advantage: $\hat{A}_{\max}(s_{t}, a_{t}) = Q_{\theta}(s_{t}, a_{t}) - \max_{j=1}^{m} Q_{\theta}(s_{t}, a_{j})$
    \end{itemize}
    where $a_{j} \sim \pi_{\phi}(\cdot \mid s_{t})$. We perform experiments with both in Section~\ref{sec:ExperimentsResults}. Additionally, we provide a batch size sensitivity analysis for CRR in Figure \ref{fig:BatchSizeCRR} : the key message is that training is faster with a larger batch size. This results does not modify our conclusions Section~\ref{sec:continuousExp}.
    \item \textbf{Conservative Q-Learning (CQL):} \\
    We limited at 500k timesteps the runs on the \texttt{Humanoid-v2} environment as it is sufficient to clearly conclude its failing. We used the recommended MuJoCo hyper-parameters from the paper. The dual gradient ascent trick to tune the weight of the Q-value regulariser $\alpha$ in the policy evaluation step didn't bring about better results. We believe the robustness of CQL is largely affected by this hyper-parameter and may therefore be the reason why CQL fails in our high dimensional continuous tasks.
\end{itemize}

The different methods were implemented using TensorFlow and Pytorch in the discrete and continuous setting respectively.

\subsection{Evaluation Protocol}

\begin{figure}[t!]
    \begin{center}
        \includegraphics[width=\textwidth]{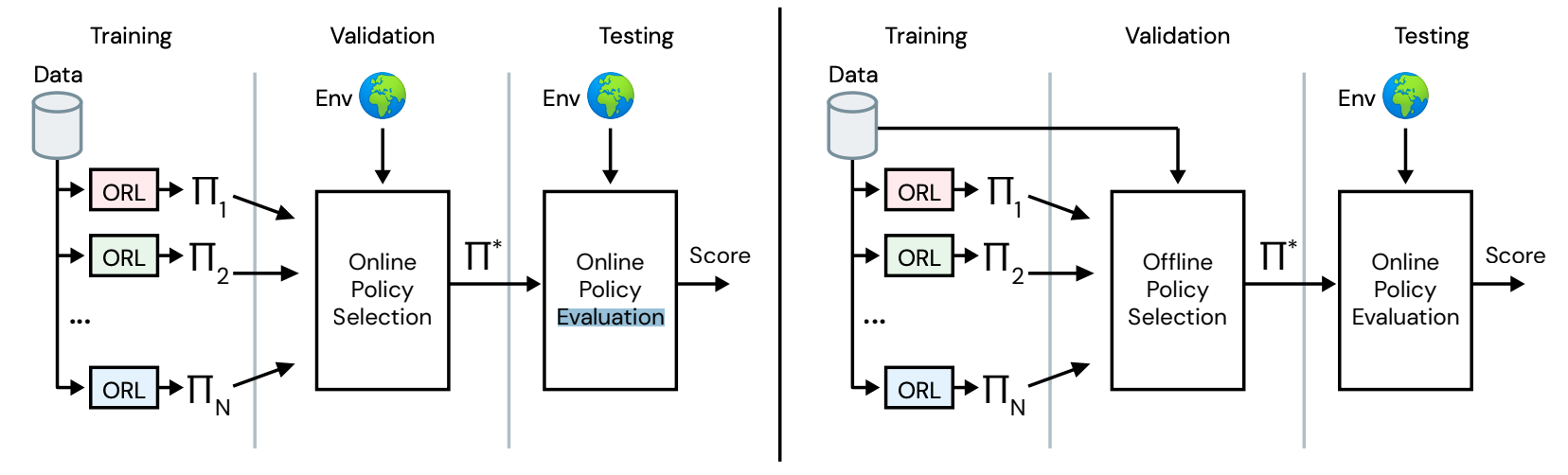} 
    \end{center}
    \captionsetup{justification=justified,margin=0cm}
    \caption{Comparison of evaluation protocols taken from \citep{Gulcehre2020RLUB}. (left) online policy selection - gives overly optimistic results as it allows perfect policy selection. (right) offline policy selection - closest to a real world use case where it is too expensive to query the environment.}
    \label{fig:evalProtocols}
\end{figure}

Within the scope of this work, we make a simplifying assumption that we have access to the environment for evaluation. It is important to note that this is usually not the case in practice, since the very motivation for doing offline reinforcement learning is that interacting with the environment is expensive or unsafe. The key difference is highlighted in the Figure \ref{fig:evalProtocols}, which shows the two most common evaluation protocols in offline reinforcement learning literature.

Due to the additional complexity that Offline Policy Evaluation (OPE) methods entail, we decided to use an online policy evaluation protocol within this work. More specifically, the performance is measured as the average trajectory return across the workers every 5000 steps. Our figures show the mean and the variance over 3 random seeds for each run.

\subsection{Hyper-parameter Choices}

\begin{table}[!ht]
\centering
    \begin{tabular}{ c | c } 
        \hline
        Hyper-parameter & Value  \\ 
        \hline
        RL batch size & 256 \\ 
        Discount factor & 0.99 \\
        Reward scaling & 1 \\
        Replay buffer size & $10^6$ \\
        Actor hidden layers & [256, 256] \\
        Actor hidden activation & ReLu \\
        Actor learning rate & $3 \cdot 10^{-4}$ \\
        Critic hidden layers & [256, 256] \\
        Critic hidden activation & ReLu \\
        Critic learning rate & $3 \cdot 10^{-4}$ \\
        Soft target $\tau$ & $5 \cdot 10^{-2}$ \\
        Offline evaluation frequency & $5 \cdot 10^{3}$ \\
        \hline
    \end{tabular}
    \caption{Default SAC hyper-parameters used for continuous control tasks.}
    \label{table:sac-params}
\end{table}
CRR mostly relies on default SAC hyper-parameters shown in table \ref{table:sac-params}. However, CQL introduces and modifies heavily the hyper-parameters, notably the policy learning rate ($3 \cdot 10^{-5}$) and the weight coefficient $\alpha$ of the Q-value regulariser.

\subsection{Environment-Specific Details}
\label{app:environments}

\begin{table}[H]
\centering
    \begin{tabular}{ c | c | c }
        \textbf{Environments}  &  \textbf{Observation space dimension}  &  \textbf{Number of actions} \\ 
        \hline 
        MiniGrid-Empty-Random-6x6-v0 & 108 & 3 \\
        MiniGrid-DistShift1-v0 & 189 & 3 \\
        \hline
    \end{tabular}
\caption{Environments with discrete observation and action spaces}
\end{table}

\begin{table}[H]
\centering
    \begin{tabular}{ c | c | c }
        \textbf{Environments}  &  \textbf{Observation space dimension}  &  \textbf{Action space dimension} \\ 
        \hline 
        PointMaze-v0 & 2 & 2 \\
        HalfCheetah-v2 & 17 & 6 \\
        Humanoid-v2 & 376 & 17  \\
        \hline
    \end{tabular}
\caption{Environments with continuous action spaces.}
\end{table}

\textbf{Specific Details about the PointMaze Environment}\\
\begin{minipage}{0.5\linewidth}
PointMaze is a simple 2D maze where the agent controls a 2D material point initialized randomly in $[- 0.1, 0.1] \times [- 0.1, - 0.7]$. The goal is to exit from the three corridors maze as fast as possible. The states correspond to the agent's position $(x_{t}, y_{t}) \in [-1, 1]$ at time $t$. The two continuous actions are mini-step increments: ($\delta x$, $\delta y$) $\in$ [$- 0.1$, $0.1$]. The episode ends once the agent reaches the exit square or the If the length of an episode exceeds 200 time steps (unsuccessful episode). At last, the reward is computed as: $r_{t} = - (x_{t} - x_{goal})^{2}  - (y_{t} - y_{goal})^{2}$.
\end{minipage}
\hfill
\begin{minipage}{0.45\linewidth}
    \begin{center}
        \includegraphics[width=0.85\textwidth]{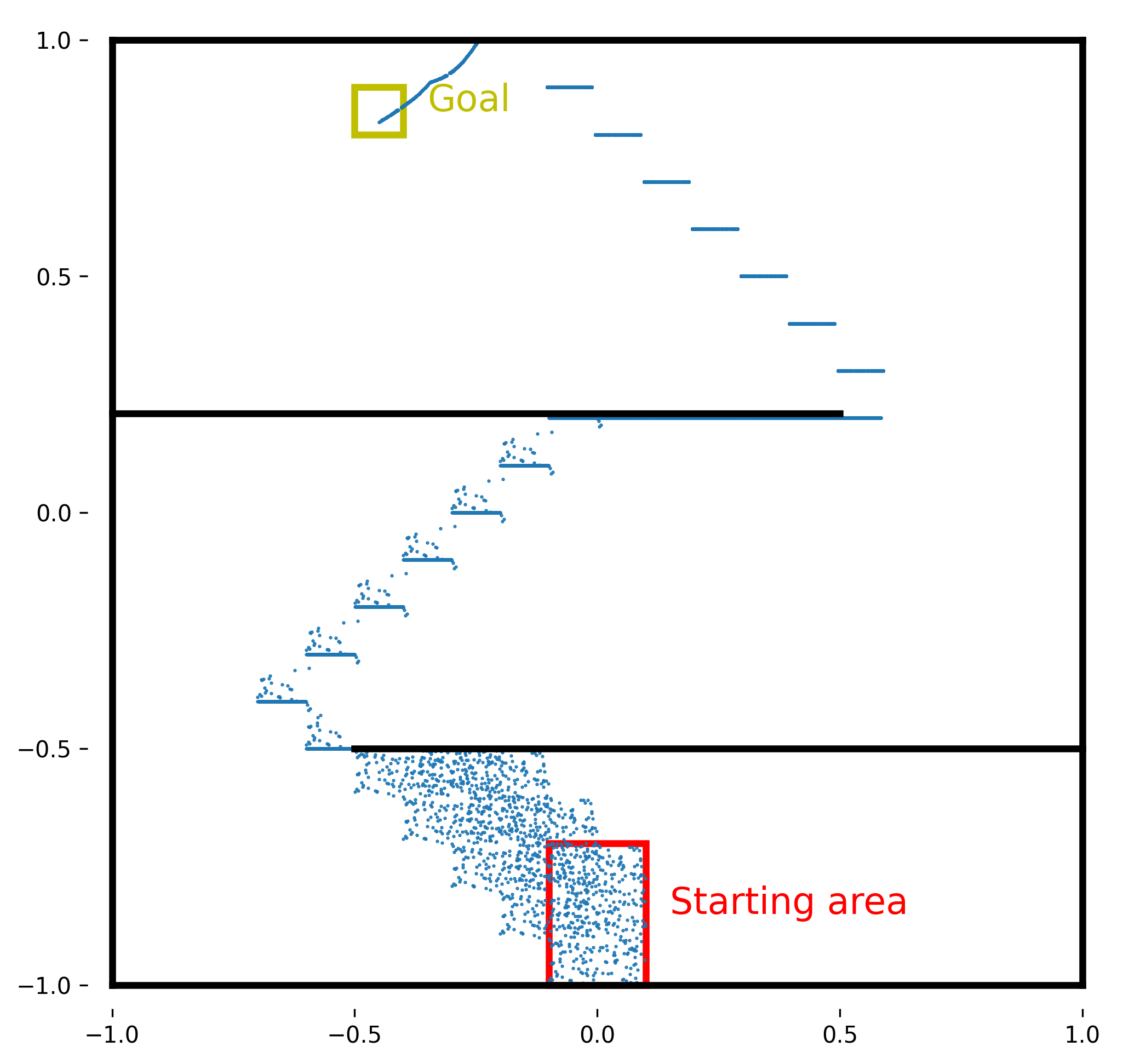}
        \captionsetup{justification=justified,margin=0cm}
        \captionof{figure}{\texttt{PointMaze-v0} environment and $10^4$ random states samples from the expert dataset}
    \end{center}
\end{minipage}

\section{Dataset Visualisation} \label{sec:DatasetVis}
In this section, we present the dataset characteristics for the discrete action spaces experiments from Section~\ref{sec:ExploringDatasetEdgeCases}.
Each $3 \times 3$ figure consists of heatmaps and histograms organised as follows: the heatmaps at the center of each side are state counts conditional on the agent facing the direction given by the position of the plot in the figure. The middle heatmap is a sum of all the directions. The total count of transitions in the dataset is in the title of each \emph{combined} heatmap (center). Histograms are placed in the corners: distribution of rewards (Upper LHS), distribution of \emph{positive} rewards (Upper RHS), episode length distribution (Lower LHS) and the proportion of actions within the dataset (Lower RHS).

The state-action coverage conditional on each facing direction can be assessed using those heatmaps: green/browning squares are for highly visited states whilst the lavender colour is for states sparsely visited.

\begin{figure}[H]
    \begin{center}
        \includegraphics[width=\textwidth]{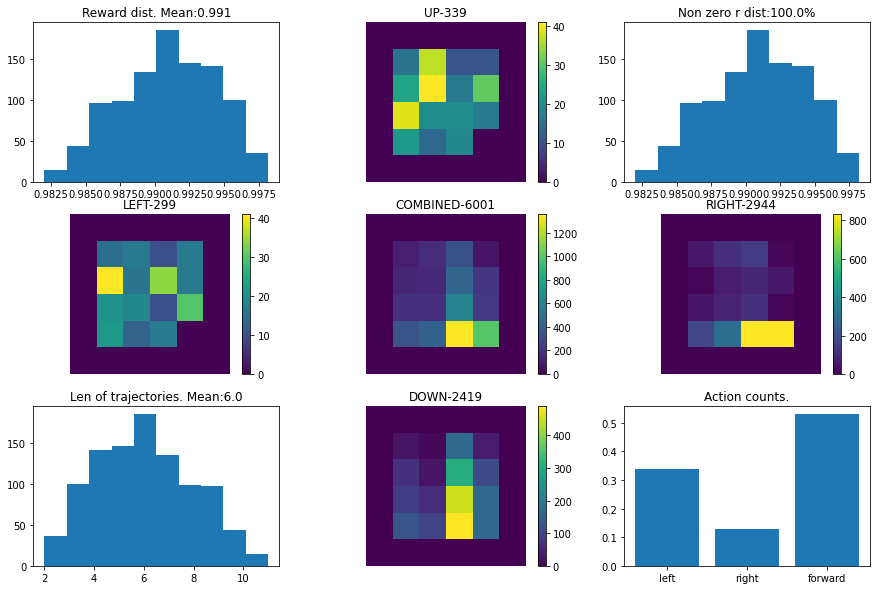}
    \end{center}
    \captionsetup{justification=justified,margin=0cm}
    \caption{Expert dataset with with 1000 episodes and $\epsilon=0$.}
    \label{fig:expertEps0}
\end{figure}

\begin{figure}[H]
    \begin{center}
        \includegraphics[width=\textwidth]{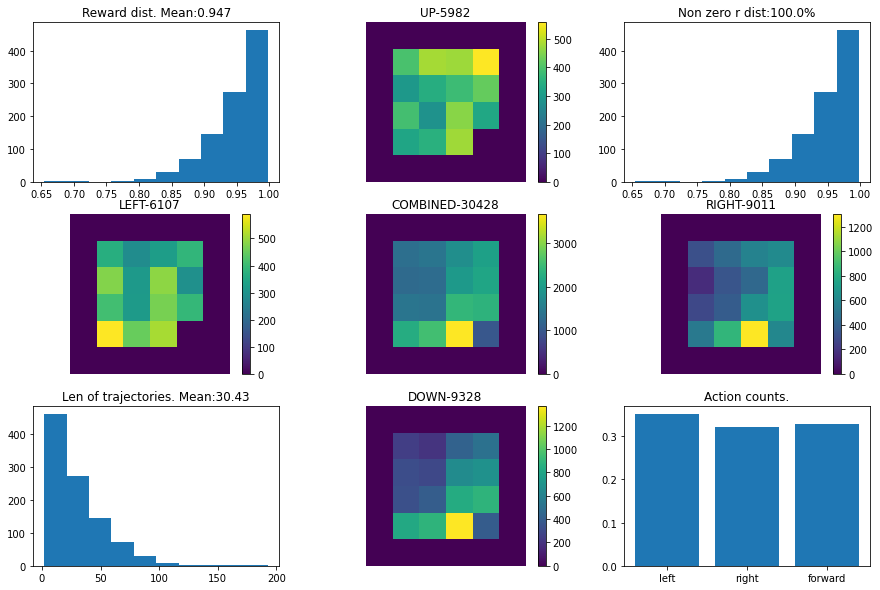}
    \end{center}
    \captionsetup{justification=justified,margin=0cm}
    \caption{Expert dataset with with 1000 episodes and $\epsilon=0.8$.}
    \label{fig:expertEps08}
\end{figure}

\begin{figure}[H]
    \begin{center}
        \includegraphics[width=\textwidth]{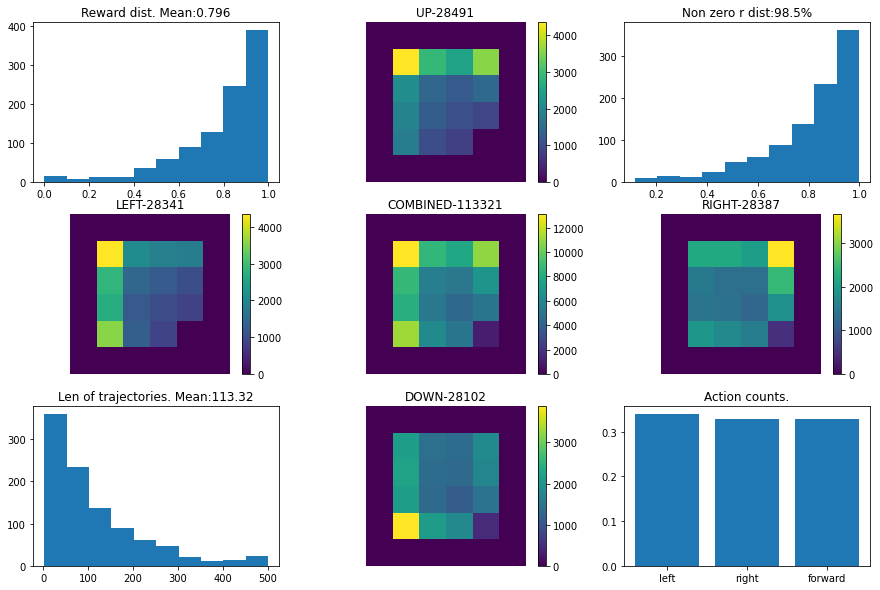}
    \end{center}
    \captionsetup{justification=justified,margin=0cm}
    \caption{Expert dataset with with 1000 episodes and $\epsilon=1$.}
    \label{fig:expertEps1}
\end{figure}

By comparing the heatmaps and action count histogram of Figures \ref{fig:expertEps0}, \ref{fig:expertEps08} and \ref{fig:expertEps1}, we notice that the state coverage gets higher across all facing directions and the action counts is closer a uniform distribution, the mean reward goes down from $0.991$ to $0.796$. Also, the episode length increases with epsilon: we can see that while the expert dataset only contains 6k transitions, the fully random dataset has more than 113k for 1,000 episodes.

\subsection{EXPERIMENT 1}

\subsection{EXPERIMENT 2}

\begin{figure}[H]
    \begin{center}
        \includegraphics[width=\textwidth]{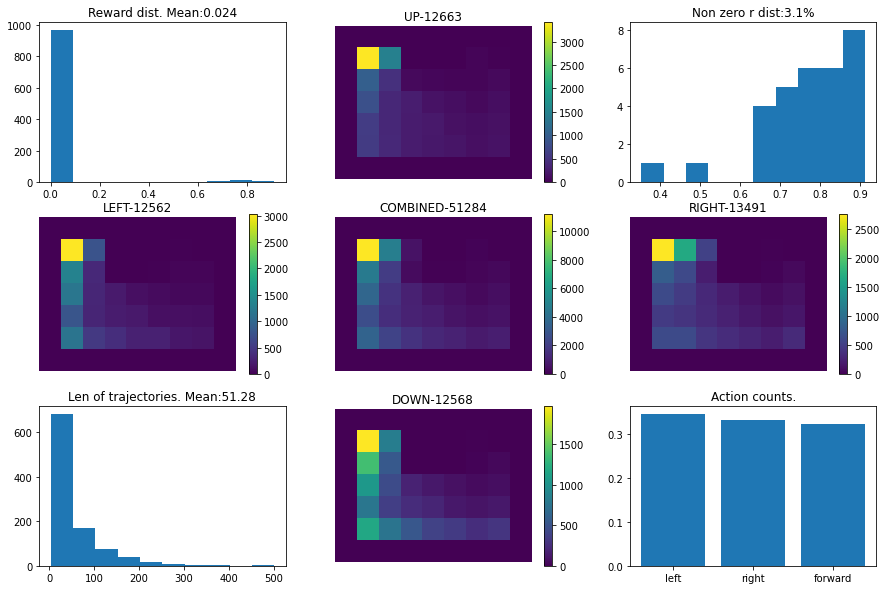}
    \end{center}
    \captionsetup{justification=justified,margin=0cm}
    \caption{Dataset collected by a random agent acting in the Lava gird world.}
    \label{fig:randomLavaVis}
\end{figure}

\subsection{EXPERIMENT 3}

\begin{figure}[H]
    \begin{center}
        \includegraphics[width=\textwidth]{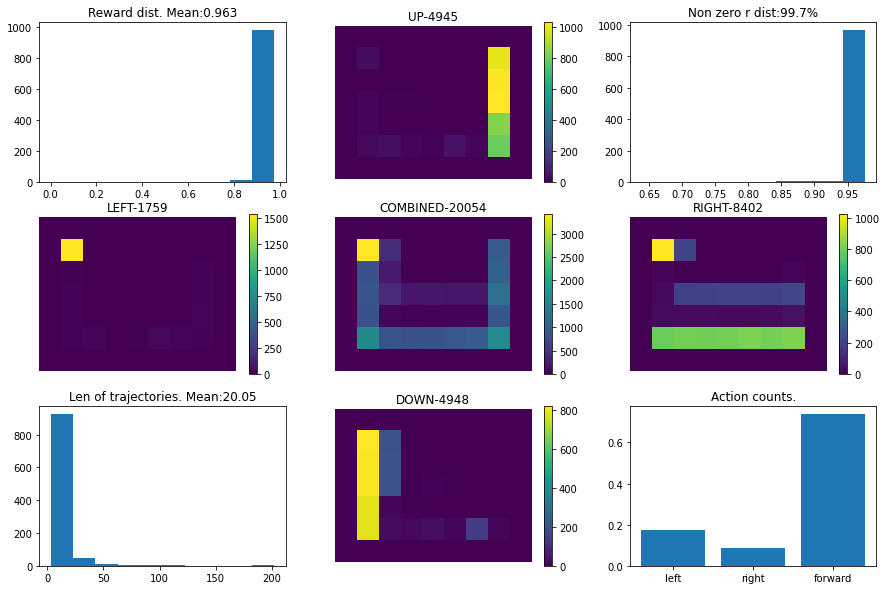}
    \end{center}
    \captionsetup{justification=justified,margin=0cm}
    \caption{Multi-modal dataset with with 1000 episodes as described in \ref{app:discreteProtocol}}
    \label{fig:MultiModalEVis}
\end{figure}

\newpage
\clearpage

\section{Supplemental results} \label{sec:MisResults}
\subsection{Hyperparameter search}
Experiments below are ran on datasets collected by an $\epsilon$-greedy expert in the Empty-Random MiniGrid world as described in Appendix \ref{app:discreteProtocol}.

\begin{table}[H]
    \centering 
    \begin{tabular}{ |c||c|c| }
         \hline
         \textbf{Epsilon}     & avg. reward & avg. ep. length \\
         \hline
         0     & 0.991 & 6 \\
         0.3   & 0.986 & 8.54 \\
         0.6   & 0.974 & 15.52 \\
         0.8   & 0.947 & 30.43 \\
         0.9   & 0.908 & 51.86 \\
         1     & 0.796 & 113.32 \\
         \hline
    \end{tabular}
    \captionsetup{justification=justified,margin=0cm}
    \caption{Summary of the key characteristics of each dataset.}
    \label{tab:DataCharTable}
\end{table}
        
\begin{figure}[H]
    \begin{center}
        \includegraphics[width=0.49\textwidth]{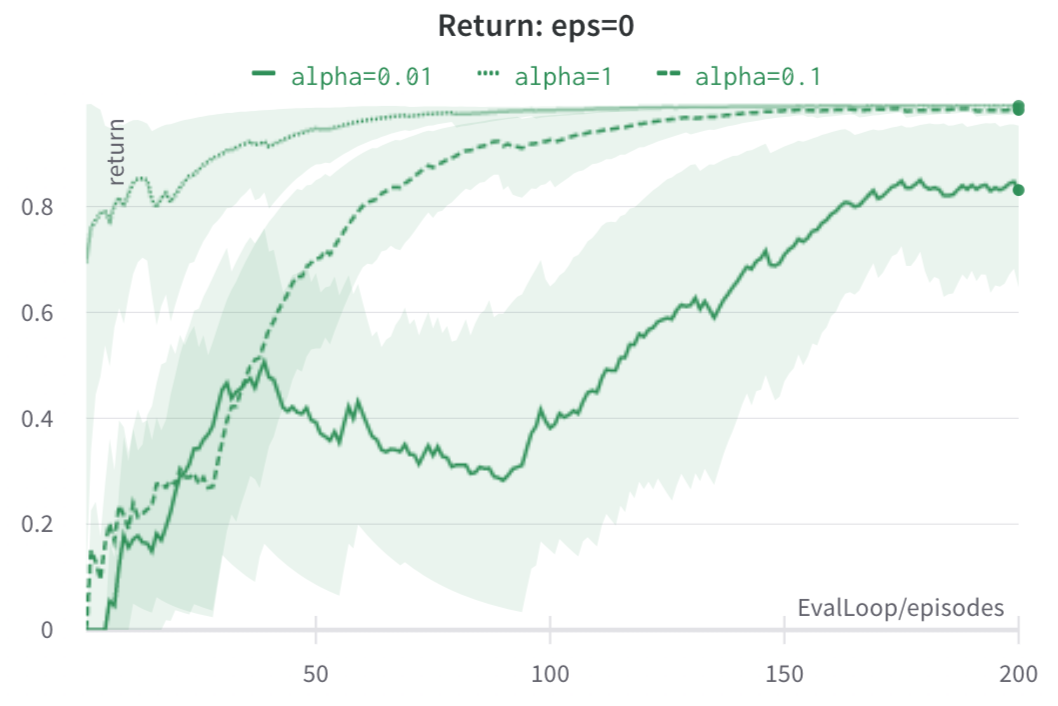} \ 
        \includegraphics[width=0.49\textwidth]{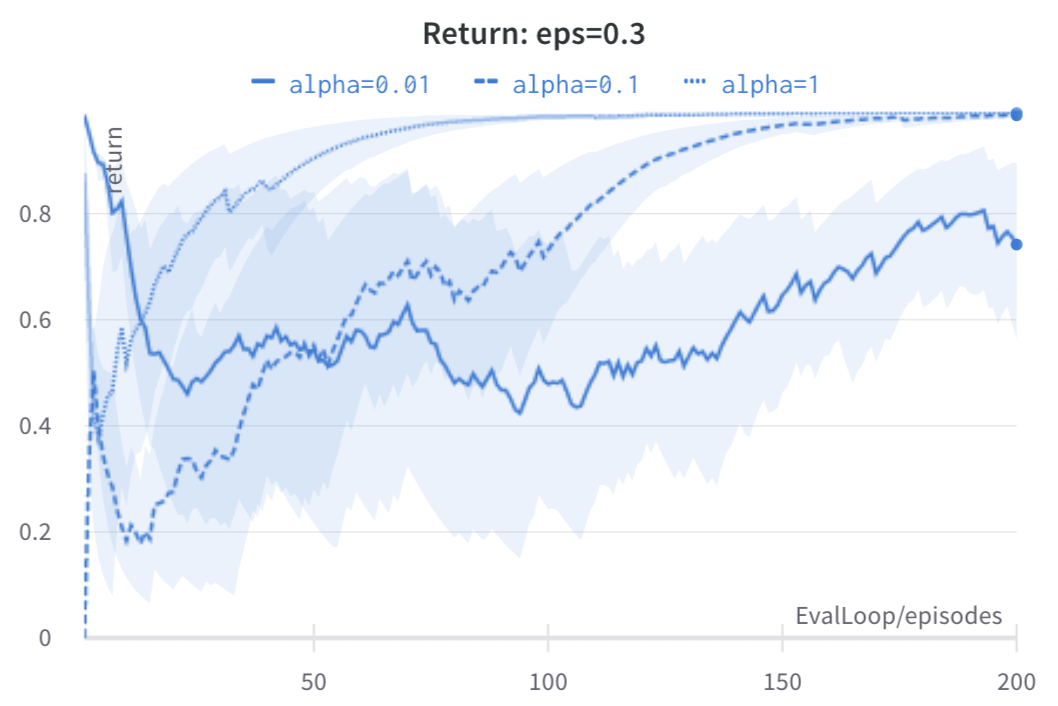} \\
        \includegraphics[width=0.49\textwidth]{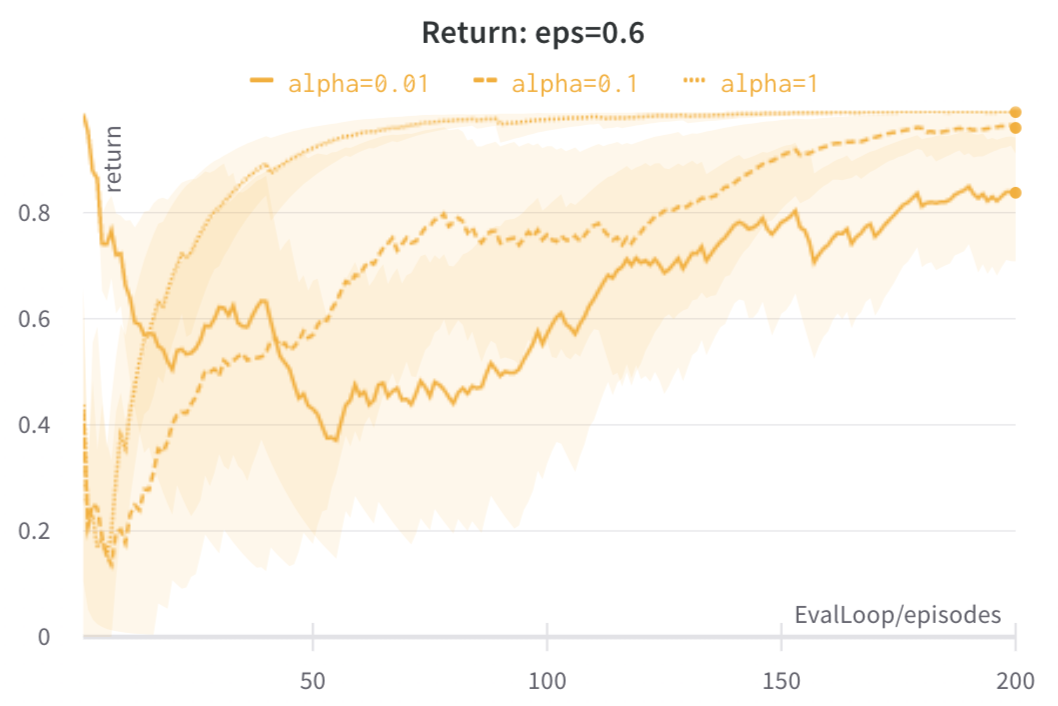} \
        \includegraphics[width=0.49\textwidth]{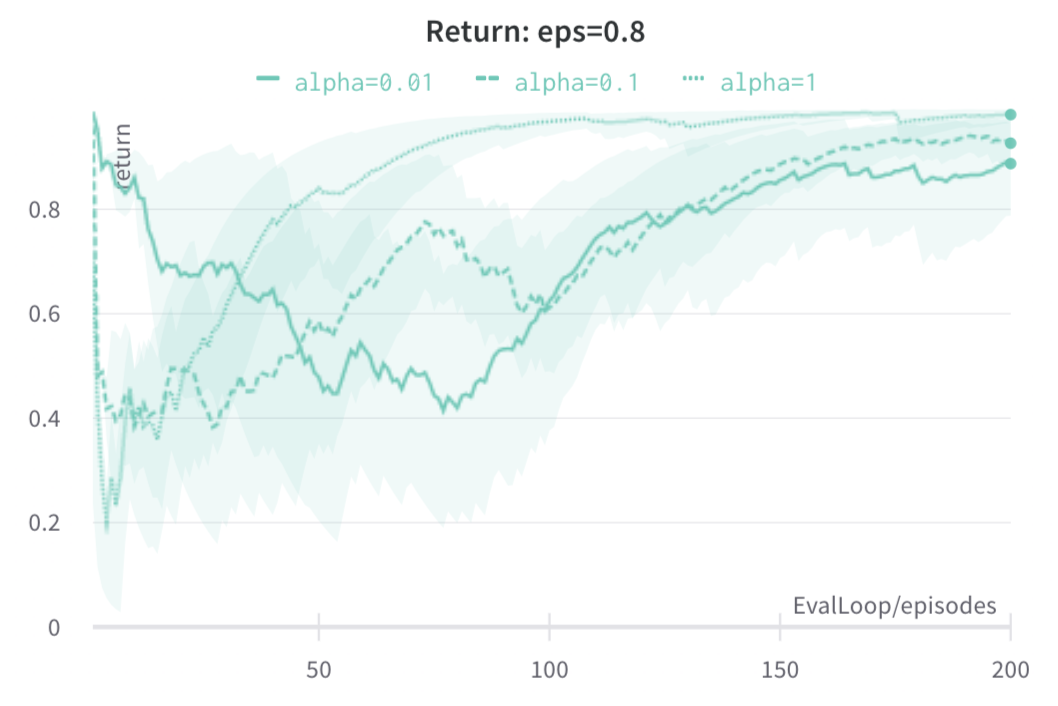} \\
        \includegraphics[width=0.49\textwidth]{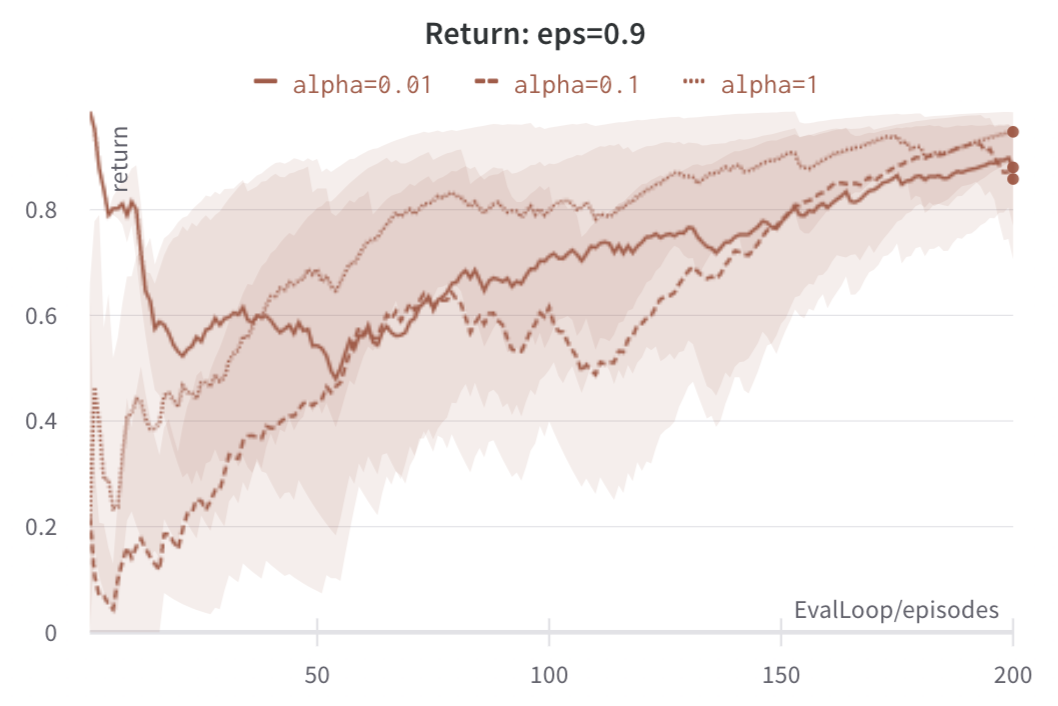} \
        \includegraphics[width=0.49\textwidth]{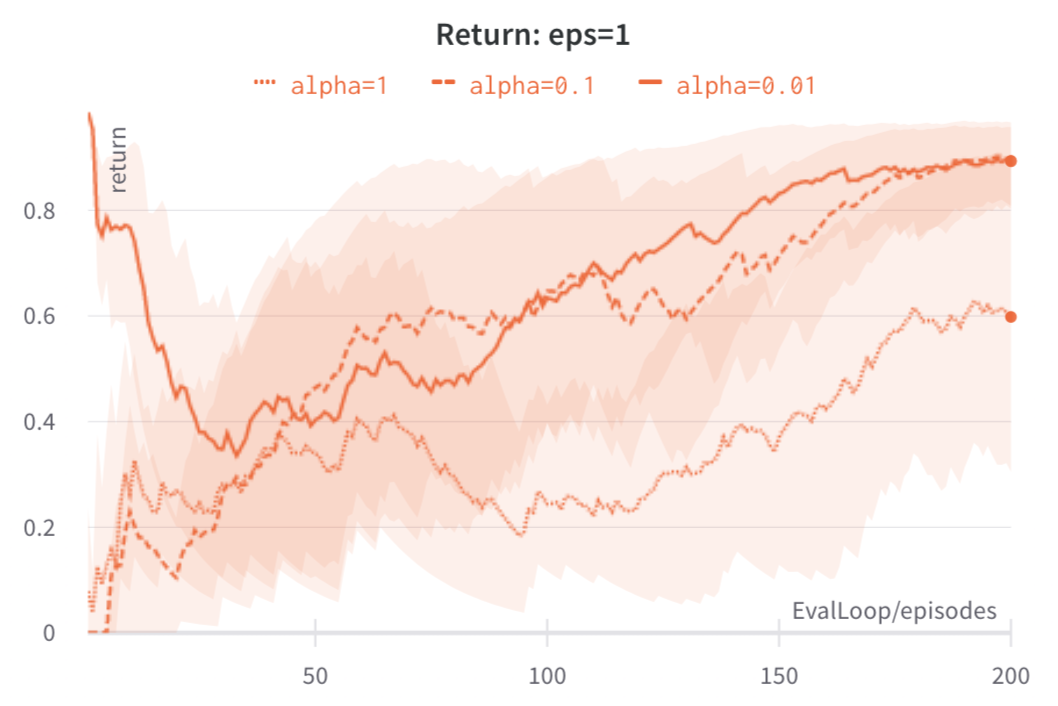}
    \end{center}
    \captionsetup{justification=justified,margin=0cm}
    \caption{Performance of CQL as a function of the dataset quality.}
    \label{fig:CQLexperiment}
\end{figure}

\begin{figure}[H]
    \begin{center}
        \includegraphics[width=0.49\textwidth]{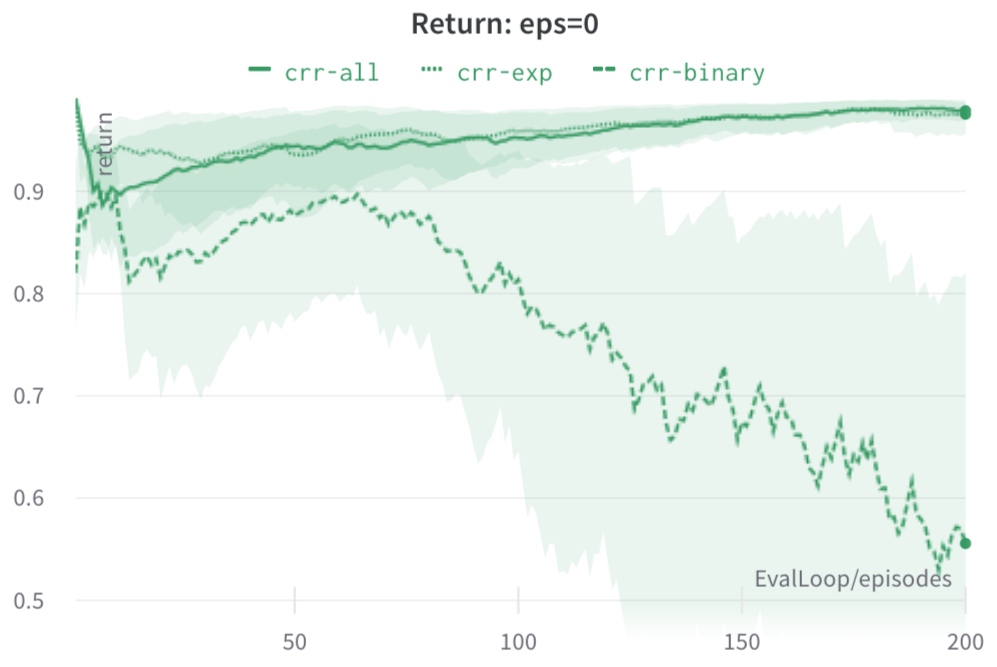} \ 
        \includegraphics[width=0.49\textwidth]{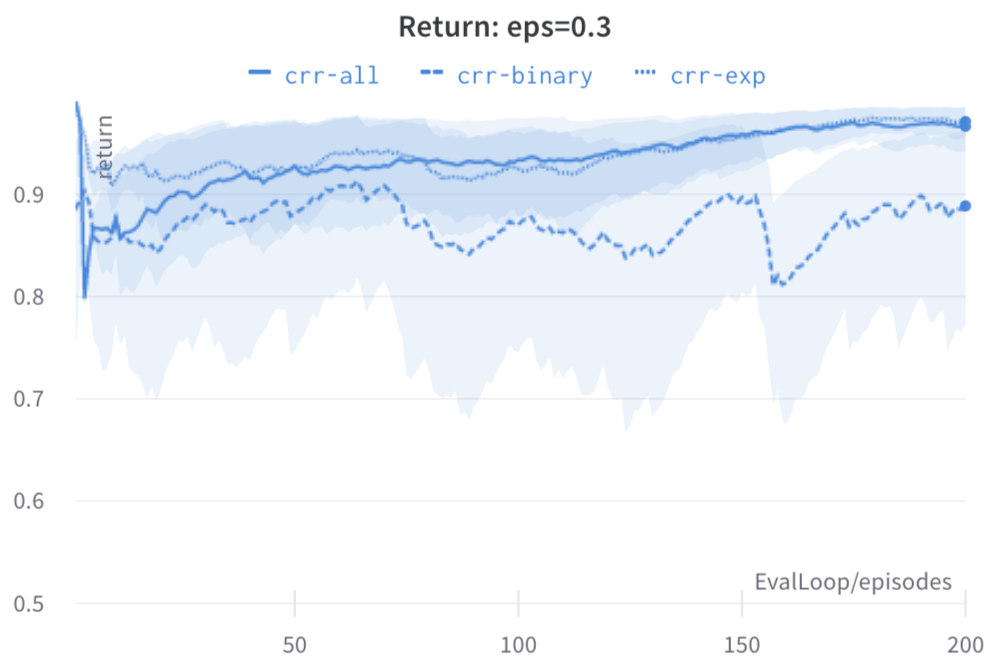} \\
        \includegraphics[width=0.49\textwidth]{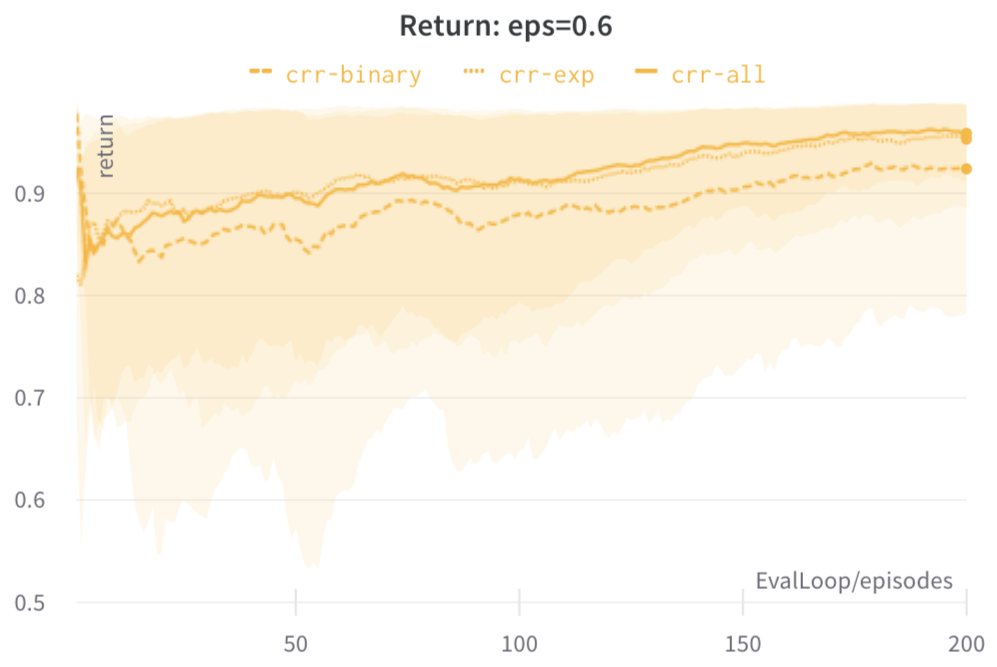} \
        \includegraphics[width=0.49\textwidth]{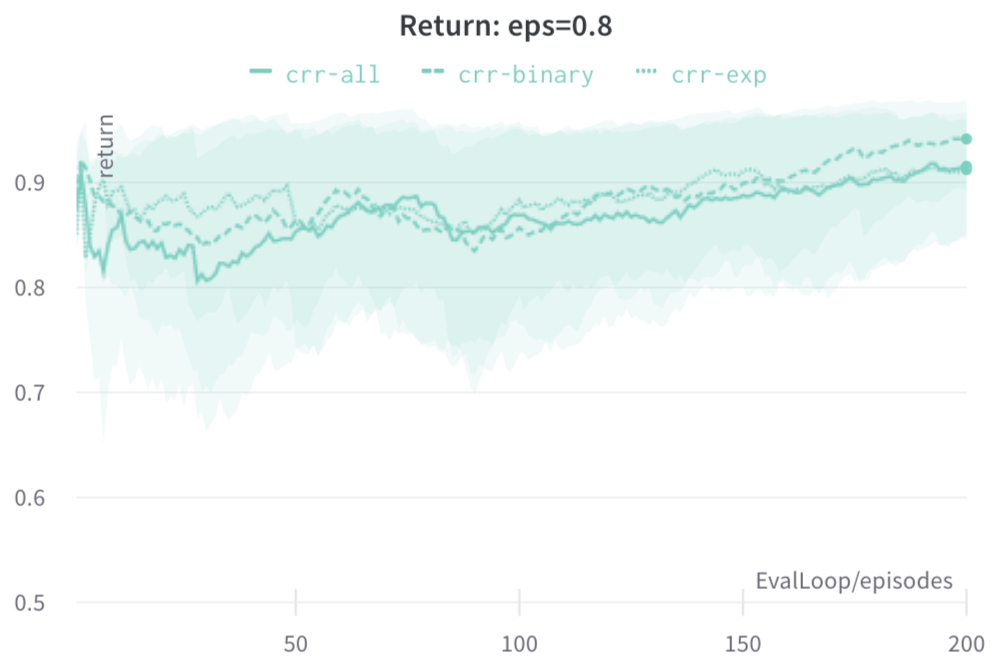} \\
        \includegraphics[width=0.49\textwidth]{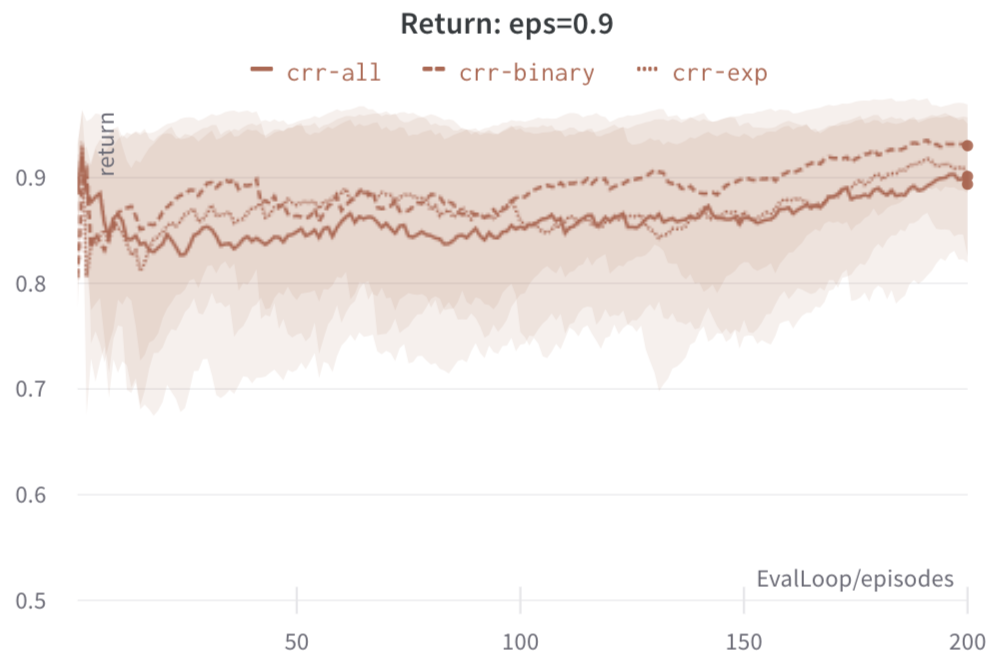} \
        \includegraphics[width=0.49\textwidth]{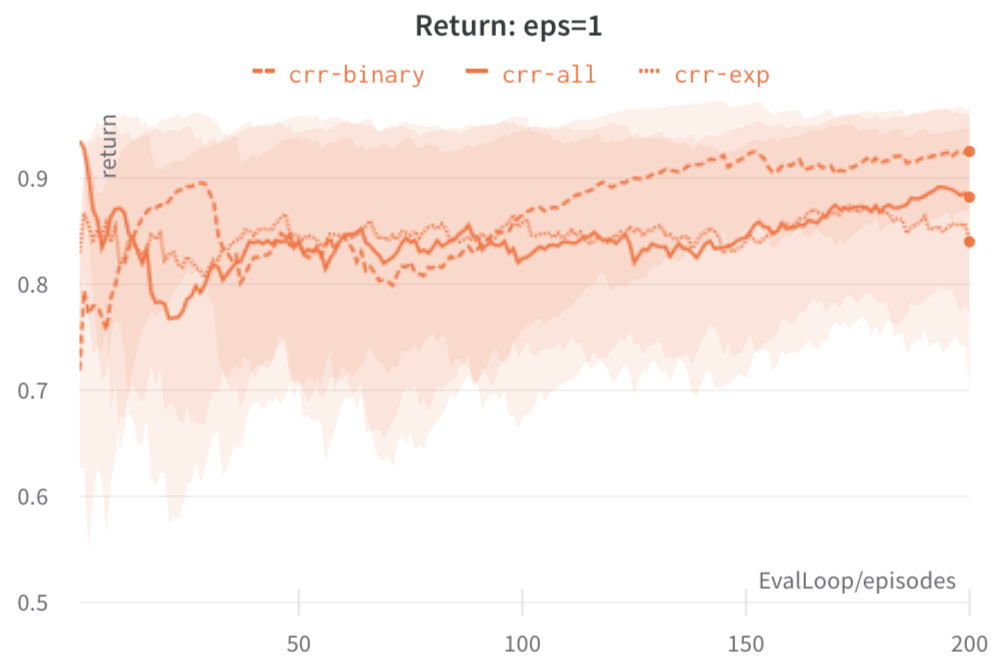}
    \end{center}
    \captionsetup{justification=justified,margin=0cm}
    \caption{Performance of CRR as a function of the dataset quality.}
    \label{fig:CRRexperiment}
\end{figure}

\section{Online fine-tuning study}
\paragraph{I have access to a simulator, is it possible to get further gains of performance with fine-tuning?}

\begin{figure}[h!]
    \begin{center}
        \includegraphics[width=\textwidth]{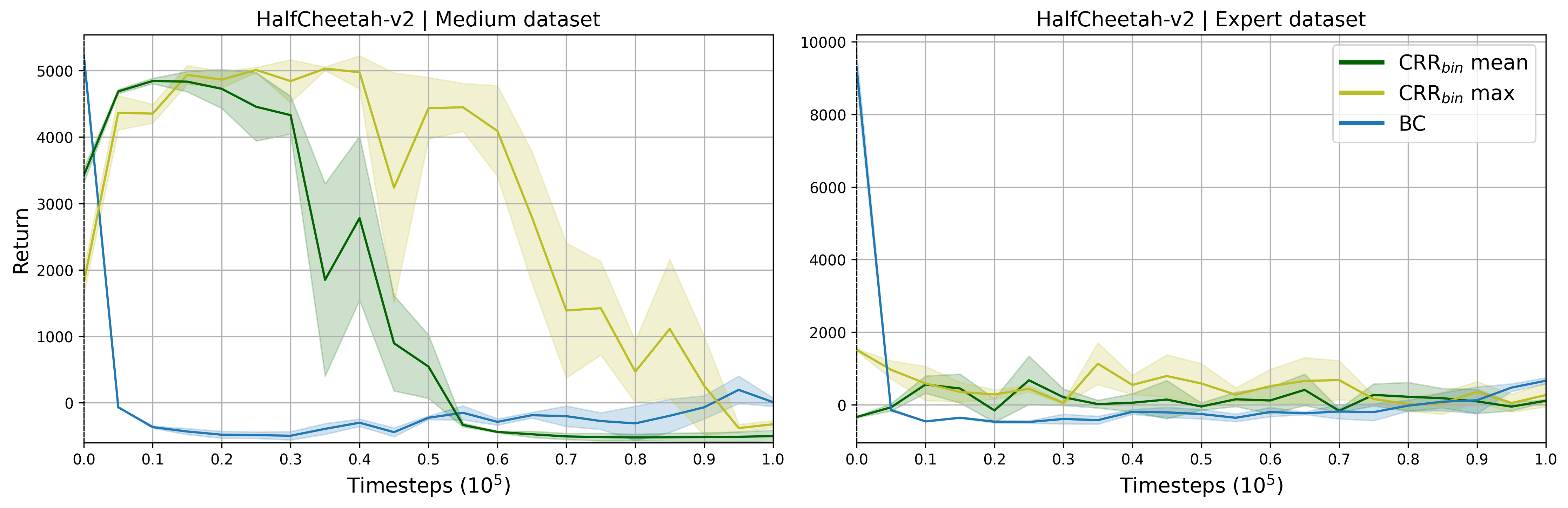}
    \end{center}
    \captionsetup{justification=justified,margin=0cm}
    \caption{Fine-tuning for $10^5$ timesteps after $10^6$ fully offline training steps on \texttt{HalfCheetah-v2}.}
    \label{fig:FineTuning}
\end{figure}

Sometimes in real-world scenarios, environment simulators can be accessible but they can be computationally costly and/or time-consuming. Nevertheless, after purely offline training, the policy can be deployed in a simulated environment to perform a limited extra exploration which we refer as \emph{fine-tuning}. This technique is investigated as a way to boost the agent performance with a small number of online trial-and-error interactions \citep{Nair2020AcceleratingOR}. We first train offline an agent for $10^6$ timesteps then fine-tune the policy for $10^5$ online interactions in the simulator (see Figure~\ref{fig:FineTuning}).

In the medium quality setting, BC suffers from performance collapse before slowly going back up again. CRR$_{bin}$ performance first reaches the average return contained in the dataset. Then, the performance crumbles entirely. Combining offline RL methods like CRR (CQL is not tested here as the offline training was not successful) with fine-tuning seems to require more care and tuning and doesn't seem to be applicable out-of-the-box.

\paragraph{Further investigation on CRR: batch size analysis.}

\begin{figure}[h!]
    \begin{center}
        \includegraphics[width=\textwidth]{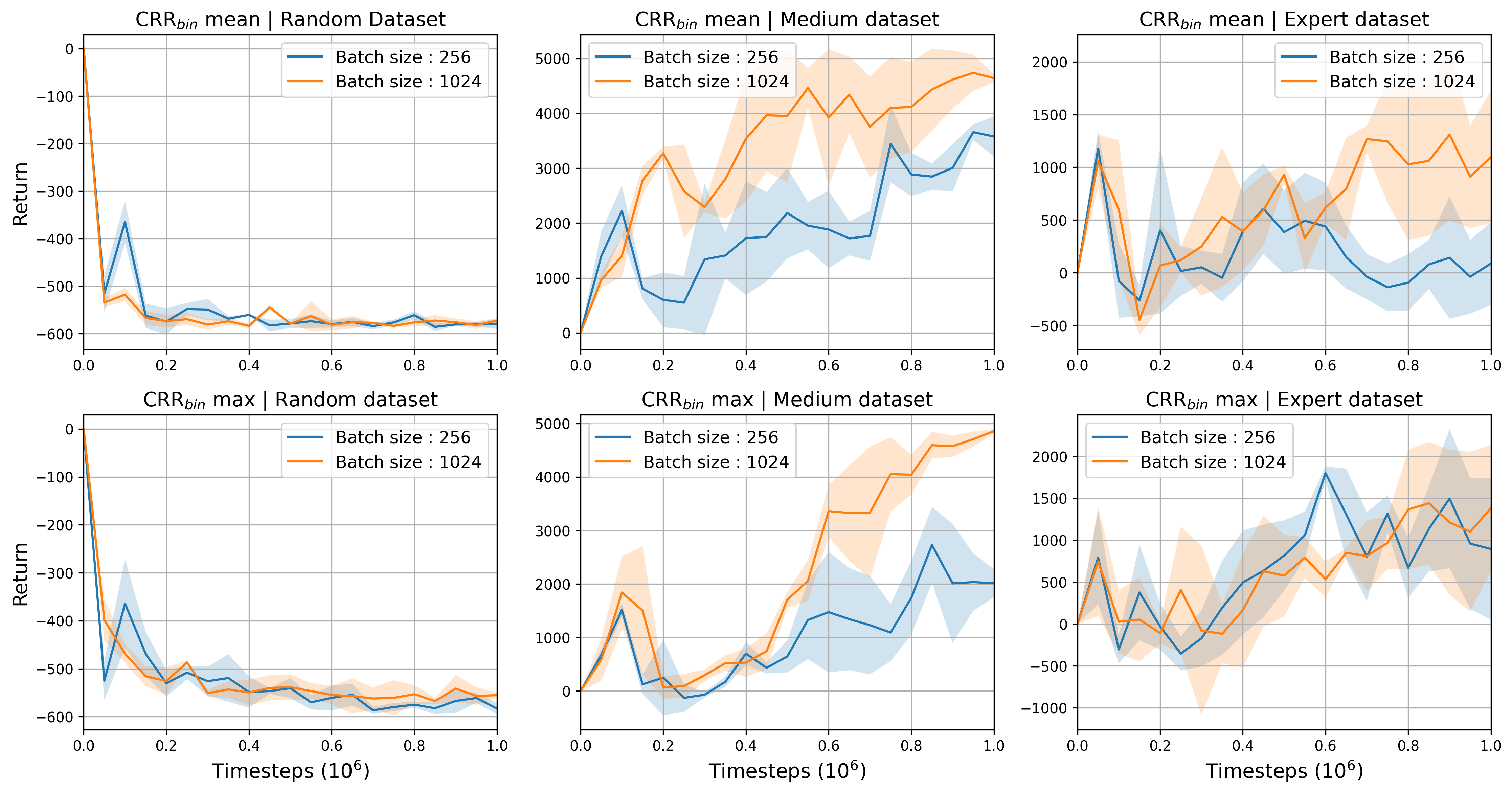}
    \end{center}
    \captionsetup{justification=justified,margin=0cm}
    \caption{Batch size sensitivity analysis for CRR on \texttt{HalfCheetah-v2}.}
    \label{fig:BatchSizeCRR}
\end{figure}

We show in Figure \ref{fig:BatchSizeCRR} that batch size hyper-parameter can play an important role for CRR, especially in the medium dataset setting. Sampling more examples simply gives a better estimate of the \emph{filtered} function and therefore the policy is more efficiently pushed towards the best actions. However, a larger batch size does not affect the conclusions in Section~\ref{sec:continuousExp}.

\end{document}